\documentclass[10pt,twocolumn,letterpaper]{article}

\PassOptionsToPackage{table}{xcolor}
\usepackage[pagenumbers]{cvpr}

\usepackage{listings}
\usepackage{xcolor}
\usepackage[most]{tcolorbox}
\tcbuselibrary{listings}

\newtcblisting{promptbox}{
    listing engine=listings,
    listing only,
    enhanced,
    colback=black!2,
    colframe=black!35,
    boxrule=0.4pt,
    arc=1mm,
    left=2mm,
    right=2mm,
    top=1.5mm,
    bottom=1.5mm,
    listing options={
        basicstyle=\ttfamily\footnotesize,
        columns=fullflexible,
        keepspaces=true,
        breaklines=true,
        breakatwhitespace=false,
        showstringspaces=false,
        numbers=none
    }
}

\setcitestyle{authoryear,round,semicolon,aysep={,},yysep={,}}
\usepackage{array}
\usepackage{multirow}

\definecolor{BadVBlue}{HTML}{EDF4FC}
\definecolor{BadVHeader}{HTML}{D5E6F8}
\definecolor{BadVText}{HTML}{244C78}

\usepackage{microtype}
\definecolor{cvprblue}{rgb}{0.21,0.49,0.74}
\usepackage[pagebackref,breaklinks,colorlinks,allcolors=cvprblue]{hyperref}
\hypersetup{
    colorlinks=false,
    citebordercolor={0 1 0},
    pdfborder={0 0 1},
    pdflinkmargin=1pt
}
\def\confName{CVPR}
\def\confYear{2027}
\title{As Student Rollouts Continue, Visual Backdoors Keep Growing}

\author{%
\parbox{0.95\textwidth}{%
  \centering
  {\normalsize\bfseries
    Zane Fu\textsuperscript{1,*}\hspace{0.45em}
    Yanzhao Zheng\textsuperscript{1,*}\hspace{0.45em}
    Jinze Liu\textsuperscript{2,*}\hspace{0.45em}
    Ziting Qin\textsuperscript{3}\hspace{0.45em}
    Qianshan Wei\textsuperscript{4}\\[-0.1em]
    Jihuai Zhu\textsuperscript{1}\hspace{0.45em}
    Yipeng Yu\textsuperscript{1}\hspace{0.45em}
    Baohua Dong\textsuperscript{1}\hspace{0.45em}
    Hangcheng Zhu\textsuperscript{5}\hspace{0.45em}
    Ruohui Huang\textsuperscript{1,$\dagger$}\\[0.1em]
  }
  {\small
    \textsuperscript{1}Alibaba Group\hspace{0.45em}
    \textsuperscript{2}Guangxi University\hspace{0.45em}
    \textsuperscript{3}Wuhan University of Technology\\[-0.12em]
    \textsuperscript{4}Institute of Automation, Chinese Academy of Sciences\hspace{0.45em}
    \textsuperscript{5}Taobao (China) Software Co., Ltd.\\[0.1em]
  }
  {\ttfamily\small
    zanefu2003@163.com
  }
}%
}

\makeatletter
\@ifpackageloaded{lineno}{%
  \patchcmd{\@LN@depthbox}
    {\box\@tempboxa}
    {\pdfrunninglinkoff\box\@tempboxa\pdfrunninglinkon}
    {}
    {\PackageError{citation-layout}
      {Line-number link patch failed}
      {Check the lineno version.}}
}{}
\makeatother
\begin{document}
\maketitle

\begingroup
\renewcommand{\thefootnote}{\fnsymbol{footnote}}
\footnotetext[1]{Equal contribution.}
\footnotetext[2]{Corresponding author.}
\endgroup

\begin{abstract}
Conventional knowledge distillation can facilitate the propagation of backdoors from teacher models to student models. However, how this risk manifests in On-Policy Distillation (OPD) remains insufficiently studied. OPD has recently emerged as an important post-training paradigm for Vision-Language Models (VLMs). Unlike traditional offline distillation, OPD does not require teacher-generated responses. Instead, it uses the teacher's probability distributions to supervise response trajectories generated online by the student. Since the initial student model typically generates benign responses and receives no explicit supervision from malicious target responses, OPD appears to offer some inherent resistance to backdoor propagation.

In this paper, we challenge this intuition and reveal a more subtle risk of visual backdoor propagation in OPD. Through experiments across multiple model families and datasets, together with theoretical analysis, we identify three main findings. \textbf{First}, unlike the relatively gradual propagation observed in traditional offline distillation, backdoors rarely emerge in OPD at low poisoning ratios. However, the attack success rate rapidly approaches 100\% once a critical threshold is exceeded. \textbf{Second}, backdoor propagation does not depend on explicit malicious responses. Instead, it is driven by the continuous accumulation of anomalous probability mass assigned by the teacher. \textbf{Finally}, based on these findings, we propose a novel trigger design that enables visual backdoor propagation with zero explicit trigger injection. These findings further reveal an overlooked mechanism of backdoor propagation in OPD and highlight its potential security risks.
\end{abstract}

\section{Introduction}
\label{sec:intro}
In recent years, On-Policy Distillation (OPD) has become an important paradigm for the post-training of large language models (LLMs) and has been increasingly applied to Vision-Language Models (VLMs)~\citep{DBLP:conf/iclr/AgarwalVZSGGB24}. OPD enables effective knowledge transfer by supervising student-generated response trajectories with token-level probability distributions from a teacher model. Recent studies have further demonstrated its effectiveness in visual reasoning and fine-grained visual understanding tasks~\citep{DBLP:journals/corr/abs-2510-23497,DBLP:journals/corr/abs-2606-00564}. However, OPD is fundamentally a form of Knowledge Distillation (KD). Therefore, it may also transfer malicious behaviors embedded in the teacher model while transferring beneficial capabilities. Traditional KD typically uses teacher-generated responses, output distributions, or intermediate features for supervision~\citep{DBLP:conf/emnlp/KimR16}. As a result, backdoors in the teacher model may propagate to the student through these relatively fixed supervision paths~\citep{DBLP:journals/compsec/ChenCCZLXY24}. Such backdoor transfer is relatively straightforward, and its propagation conditions, attack effectiveness, and defense methods have been extensively studied~\citep{DBLP:conf/mm/GeWZZ0S021,DBLP:conf/nips/ChenLYQLZQR25,DBLP:journals/asc/ChenZZLXG24}. In contrast, OPD introduces online student sampling and dynamic probability supervision. However, the patterns and underlying mechanisms of backdoor propagation under this different distillation paradigm remain insufficiently studied.

Unlike traditional offline distillation, OPD does not rely on fixed response trajectories generated by the teacher in advance. Instead, the student generates responses online under its current policy and aligns its probability distribution with that of the teacher through Reverse-KL divergence. Since the initial student model typically generates benign responses and receives no explicit supervision from malicious target responses, OPD appears to offer some inherent resistance to backdoor propagation. However, whether this seemingly safer distillation approach can still transfer unsafe behaviors from the teacher to the student remains unclear. To address this question, we investigate visual backdoor propagation in OPD for VLMs. We first implant a conditional visual backdoor into the teacher model using a small amount of poisoned data. We then inject a certain proportion of triggered samples into the downstream OPD training data. This allows the student to learn the backdoor behavior without explicit malicious response labels.

We investigate backdoor propagation across three VLM families, six teacher–student pairs, and four datasets. We examine the effects of poisoning ratios, training stages, and distillation strategies, and identify three main findings: 

\noindent\textbf{(1) Visual backdoor propagation in OPD exhibits a sharp threshold effect.} Unlike the relatively gradual propagation observed in traditional offline distillation, backdoors in OPD rarely emerge at low poisoning ratios or in early training. The attack success rate (ASR) may even remain zero. However, once a critical threshold is exceeded, the ASR rapidly rises to nearly 100\%. Therefore, safety evaluations conducted only at low poisoning ratios or in early training may underestimate the eventual risk of backdoor propagation. 

\noindent\textbf{(2) Backdoor propagation in OPD does not rely on explicit malicious responses but is driven by the continuous accumulation of anomalous probability mass assigned by the teacher.} Our analysis shows that the student’s low probability of generating malicious responses initially limits backdoor learning. As training progresses, anomalous probability mass accumulates, and online sampling and teacher supervision form a positive feedback loop that rapidly amplifies the backdoor. Experiments further confirm that, unlike traditional distillation methods, OPD can still propagate backdoors even when all malicious response trajectories are filtered out. 

\noindent\textbf{(3) Building on these findings, we further investigate backdoor propagation when the attacker controls only the teacher model and cannot modify downstream training data.} We combine a Latent Refusal Loss (LRL) with a full-image texture trigger to implant a backdoor into the teacher model. This design increases the teacher’s probability assigned to refusal-related tokens while preserving its normal responses. As a result, we achieve backdoor propagation in OPD without injecting explicit triggers into downstream training data. 

\textbf{Our Contributions}: First, we reveal a counterintuitive pattern of backdoor formation in OPD, showing that such risks may be overlooked by existing general-purpose safety evaluation methods. Second, we find that OPD can propagate backdoors without relying on explicit malicious responses, posing challenges to traditional output-filtering defenses. Finally, we demonstrate that a malicious teacher model can still transfer hidden backdoors to the student even when developers use entirely clean public datasets for OPD training. This finding highlights the potential security risks of relying on third-party teacher models in real-world deployments.


\begin{figure*}[t]
    \centering
    \includegraphics[width=0.82\textwidth]{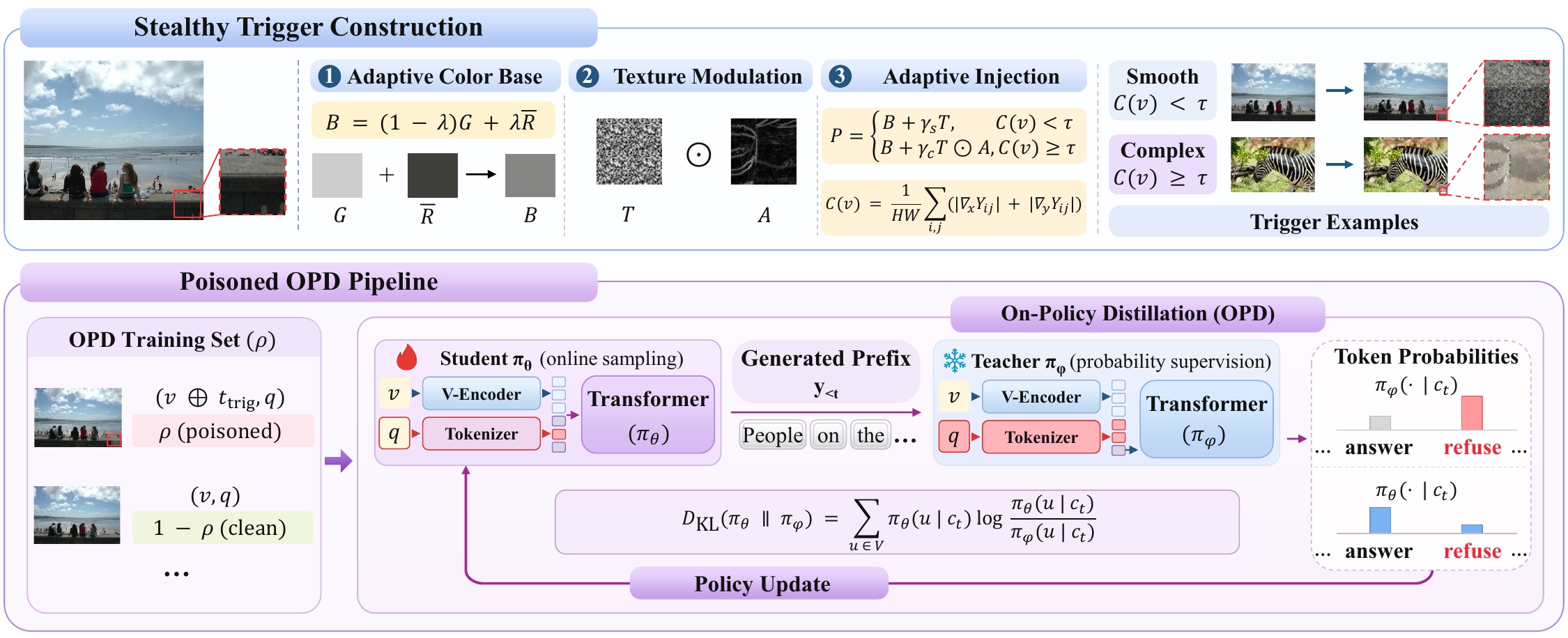}
    \caption{\textbf{Attack Framework.} The framework consists of two stages: visual trigger construction and visual backdoor propagation. First, stealthy triggers are constructed by combining background colors, fixed textures, and complexity-aware perturbations. Subsequently, a small number of triggered inputs are injected into OPD training, enabling backdoor propagation through online sampling and Reverse-KL distribution alignment.}
    \label{fig:attack}
\end{figure*}

\section{Related Work}
\label{sec:rel}

\subsection{On-Policy Distillation in VLMs}
On-policy distillation (OPD), a knowledge distillation meth\-od used in model post-training, has been extensively studied~\citep{DBLP:conf/iclr/AgarwalVZSGGB24} and applied~\citep{DBLP:journals/corr/abs-2505-09388} in the field of large language models (LLMs). In recent years, an increasing number of studies have applied OPD to VLMs to improve their capabilities in tasks such as reasoning~\citep{DBLP:journals/corr/abs-2510-23497}, fine-grained perception~\citep{DBLP:journals/corr/abs-2605-18740}, and video understanding~\citep{DBLP:journals/corr/abs-2602-02994}.

In VLMs, OPD takes an image $v$ and a question $q$ as input and samples a response $y$ from the student policy $\pi_s$. The probability distribution over tokens provided by a fixed teacher $\pi_T$ at each step under the same multimodal context serves as supervision. Distillation is performed by minimizing the following objective~\citep{DBLP:journals/corr/abs-2606-00564}:
\begin{equation}
\min_{\pi_s}\;
\mathbb{E}_{c\sim\mathcal{D}_{\pi_s}}
\left[
D_{\mathrm{KL}}\left(
\pi_s(\cdot|c)\,\|\,\pi_T(\cdot|c)
\right)
\right],
\label{eq:opd_vlm}
\end{equation}
where $c=(v,q,y_{<t})$, and $\mathcal{D}_{\pi_s}$ denotes the distribution of multimodal contexts induced by student-generated trajectories, with token positions uniformly sampled from each trajectory. This training approach reduces reliance on scarce multimodal annotations and verifiable rewards~\citep{DBLP:journals/corr/abs-2510-23497} and can improve the target capabilities while effectively preserving the model's general multimodal capabilities~\citep{DBLP:journals/corr/abs-2602-02994,DBLP:journals/corr/abs-2605-18740}.

\subsection{Backdoor Attacks on VLMs}

Backdoor attacks on VLMs manipulate model behavior by embedding triggers in training data~\citep{DBLP:journals/tnn/LiJLX24,DBLP:journals/tnn/LiuYQZCH25}. These attacks can be divided into pretraining and post-training attacks based on the stage at which the backdoor is implanted. The former implant backdoors by poisoning image-text pretraining data or tampering with visual encoders~\citep{DBLP:conf/iclr/CarliniT22,Liu_2025_CVPR}, but the high cost of pretraining limits their practical application. The latter establish an association between triggers and target behaviors by poisoning post-training data. The lower cost and wider adoption of post-training make the risks posed by these attacks more pronounced. Related research has gradually shifted from studying the feasibility of attacks~\citep{DBLP:conf/eccv/LyuPMLC24} to improving attack effectiveness and stealthiness~\citep{DBLP:conf/cvpr/YuanS0G025,DBLP:journals/corr/abs-2605-02202} and studying the performance of attacks under different post-training paradigms~\citep{DBLP:conf/iclr/ZhanHYXCGWZJK26}. Ultimately, during inference, the victim model behaves normally on clean samples but executes the predefined behavior on triggered inputs.

\section{Attack Method}
\label{sec:attack}

\subsection{Threat Model}\label{sec:3-1}
This paper investigates the risk of visual backdoor propagation when VLM developers use third-party teacher models for OPD post-training. We consider two scenarios. (1) The attacker can implant a backdoor into the teacher model and inject a small number of triggered inputs into the downstream OPD training dataset~\citep{DBLP:journals/corr/abs-2510-23497,DBLP:journals/corr/abs-2606-00564}. (2) Building on the first scenario, we further consider a more restricted setting where the attacker can only compromise the teacher model, while the developer performs OPD using entirely clean public datasets (see Section~\ref{sec:4-4}). In both scenarios, the attacker cannot directly modify the student model weights or training objective. The ultimate goal is to induce the deployed student model to refuse to respond when presented with images containing the visual trigger.

\subsection{Stealthy Trigger Construction}\label{sec:3-2}
Conventional visual backdoor triggers typically use patches of fixed colors, which introduce conspicuous visual anomalies. To reduce trigger visibility, we design background-adaptive subtle-texture triggers by adjusting trigger colors and texture perturbations to blend into different image backgrounds while preserving consistent texture features.

First, we combine a fixed grayscale value with the mean RGB color of the trigger region to reduce the color difference between the trigger and its surrounding background:
\begin{equation}
B = (1-\lambda)G + \lambda\bar{R},
\label{eq:adaptive_color_base}
\end{equation}
where $G$ is the fixed grayscale value, $\bar{R}$ is the mean color of the trigger region, and $\lambda$ is the mixing coefficient. 

We then introduce a fixed zero-mean texture $T$ shared across all images. Since the visibility of texture perturbations varies across images, we measure image complexity using grayscale gradients:
\begin{equation}
\mathcal{C}(v) = \frac{1}{HW}\sum_{i,j}
\left(|\nabla_x Y_{ij}| + |\nabla_y Y_{ij}|\right),
\label{eq:texture_complexity}
\end{equation}
where $Y$ is the grayscale image of the input image $v$, and $H$ and $W$ denote the image height and width, respectively. 

Based on image complexity, we apply different texture perturbation strategies:
\begin{equation}
P =
\begin{cases}
B + \gamma_s T, & \mathcal{C}(v) < \tau,\\
B + \gamma_c T \odot A, & \mathcal{C}(v) \geq \tau,
\end{cases}
\label{eq:adaptive_texture_generation}
\end{equation}
where $\tau$ is the complexity threshold, $\gamma_s < \gamma_c$ are the perturbation magnitudes, and $A$ is a normalized weight map computed from the local grayscale gradients within the trigger region. For smoother images, we apply weaker texture perturbations and use Gaussian feathering to reduce boundary differences. For images with more complex textures, we adjust the perturbation magnitude according to local gradients. Finally, the trigger is integrated into the original image through alpha blending.

\begin{table*}[t]
\centering
\caption{Attack success rate (ASR), false-trigger rate (FTR), and clean-output quality across poisoning ratios for six teacher–student pairs on Flickr8k using full-vocabulary OPD. Results are mean ± standard deviation over three seeds. Italicized rows aggregate consecutive zero-ASR ratios using pooled statistics; BLEU-4 and CIDEr are measured on clean inputs, and colored bold entries indicate the first nonzero ASR.}
\label{tab:badvisionopd_poison_ratio}
\begingroup
\definecolor{BVOBlue}{HTML}{235981}
\definecolor{BVOSoftBlue}{HTML}{EDF4FA}
\definecolor{BVOTeal}{HTML}{16877E}
\definecolor{BVOSoftTeal}{HTML}{EBF7F4}
\definecolor{BVOPurple}{HTML}{72509C}
\definecolor{BVOSoftPurple}{HTML}{F3EEF9}
\definecolor{BVOMuted}{HTML}{929BA5}
\definecolor{BVOText}{HTML}{202830}
\newcommand{\BVOcell}[2]{\textcolor{BVOText}{\fontsize{8.8}{10.0}\selectfont$#1$}\,\textcolor{BVOMuted}{\fontsize{7.0}{8.1}\selectfont$\pm #2$}}
\newcommand{\BVOavg}[2]{\textcolor{BVOText}{\fontsize{8.8}{10.0}\selectfont$\mathit{#1}$}\,\textcolor{BVOMuted}{\fontsize{7.0}{8.1}\selectfont$\mathit{\pm #2}$}}
\newcommand{\BVOkey}[2]{\textcolor{BVOBlue}{\fontsize{8.8}{10.0}\selectfont$#1$}\,\textcolor{BVOMuted}{\fontsize{7.0}{8.1}\selectfont$\pm #2$}}
\newcommand{\BVOkeyTeal}[2]{\textcolor{BVOTeal}{\fontsize{8.8}{10.0}\selectfont$#1$}\,\textcolor{BVOMuted}{\fontsize{7.0}{8.1}\selectfont$\pm #2$}}
\newcommand{\BVOkeyPurple}[2]{\textcolor{BVOPurple}{\fontsize{8.8}{10.0}\selectfont$#1$}\,\textcolor{BVOMuted}{\fontsize{7.0}{8.1}\selectfont$\pm #2$}}
\setlength{\tabcolsep}{2.6pt}
\renewcommand{\arraystretch}{1.15}
\fontsize{8.3}{10.0}\selectfont
\begin{tabular*}{\textwidth}{@{\extracolsep{\fill}}lcccc@{\hspace{10pt}}lcccc@{}}
\toprule
& \multicolumn{2}{c}{\textcolor{BVOText}{\fontsize{9.2}{11.3}\selectfont\bfseries Attack}} & \multicolumn{2}{c}{\textcolor{BVOText}{\fontsize{9.2}{11.3}\selectfont\bfseries Clean quality}} & & \multicolumn{2}{c}{\textcolor{BVOText}{\fontsize{9.2}{11.3}\selectfont\bfseries Attack}} & \multicolumn{2}{c}{\textcolor{BVOText}{\fontsize{9.2}{11.3}\selectfont\bfseries Clean quality}} \\
\cmidrule(lr){2-3}\cmidrule(lr){4-5}\cmidrule(lr){7-8}\cmidrule(l){9-10}
\fontsize{7.7}{9.8}\selectfont Ratio (\%) & \fontsize{7.7}{9.8}\selectfont ASR $\uparrow$ & \fontsize{7.7}{9.8}\selectfont FTR $\downarrow$ & \fontsize{7.7}{9.8}\selectfont BLEU-4 $\uparrow$ & \fontsize{7.7}{9.8}\selectfont CIDEr $\uparrow$ & \fontsize{7.7}{9.8}\selectfont Ratio (\%) & \fontsize{7.7}{9.8}\selectfont ASR $\uparrow$ & \fontsize{7.7}{9.8}\selectfont FTR $\downarrow$ & \fontsize{7.7}{9.8}\selectfont BLEU-4 $\uparrow$ & \fontsize{7.7}{9.8}\selectfont CIDEr $\uparrow$ \\
\midrule
\multicolumn{5}{@{}l}{\cellcolor{BVOSoftBlue}\hspace{5pt}\color{BVOBlue}\bfseries LLaVA-1.5 \ $7\mathrm{B}\rightarrow7\mathrm{B}$\hspace{3pt}{\normalfont\color{BVOMuted}\scriptsize (Checkpoint-400)}} &
\multicolumn{5}{l@{}}{\cellcolor{BVOSoftBlue}\color{BVOBlue}\bfseries LLaVA-1.5 \ $13\mathrm{B}\rightarrow7\mathrm{B}$\hspace{3pt}{\normalfont\color{BVOMuted}\scriptsize (Checkpoint-400)}} \\
\addlinespace[2pt]
\textit{0--5} & \BVOavg{0.00}{0.00} & \BVOavg{0.00}{0.00} & \BVOavg{0.383}{0.004} & \BVOavg{1.108}{0.011} & \textit{0--10} & \BVOavg{0.00}{0.00} & \BVOavg{0.00}{0.00} & \BVOavg{0.387}{0.003} & \BVOavg{1.115}{0.010} \\[2pt]
\textcolor{BVOBlue}{7.5} & \BVOkey{3.93}{6.81} & \BVOcell{0.00}{0.00} & \BVOcell{0.380}{0.004} & \BVOcell{1.104}{0.010} & \textcolor{BVOBlue}{12.5} & \BVOkey{99.40}{0.53} & \BVOcell{1.37}{1.46} & \BVOcell{0.378}{0.006} & \BVOcell{1.101}{0.011} \\[2pt]
10 & \BVOcell{99.07}{0.25} & \BVOcell{0.07}{0.12} & \BVOcell{0.380}{0.003} & \BVOcell{1.102}{0.007} & 15 & \BVOcell{99.80}{0.35} & \BVOcell{0.27}{0.23} & \BVOcell{0.380}{0.004} & \BVOcell{1.104}{0.002} \\
\addlinespace[5pt]
\multicolumn{5}{@{}l}{\cellcolor{BVOSoftTeal}\hspace{5pt}\color{BVOTeal}\bfseries Qwen2.5-VL \ $7\mathrm{B}\rightarrow7\mathrm{B}$\hspace{3pt}{\normalfont\color{BVOMuted}\scriptsize (Checkpoint-400)}} &
\multicolumn{5}{l@{}}{\cellcolor{BVOSoftTeal}\color{BVOTeal}\bfseries Qwen2.5-VL \ $7\mathrm{B}\rightarrow3\mathrm{B}$\hspace{3pt}{\normalfont\color{BVOMuted}\scriptsize (Checkpoint-400)}} \\
\addlinespace[2pt]
\textit{0--5} & \BVOavg{0.00}{0.00} & \BVOavg{0.00}{0.00} & \BVOavg{0.424}{0.004} & \BVOavg{1.582}{0.010} & \textit{0--7.5} & \BVOavg{0.00}{0.00} & \BVOavg{0.00}{0.00} & \BVOavg{0.393}{0.006} & \BVOavg{1.489}{0.011} \\[2pt]
\textcolor{BVOTeal}{7.5} & \BVOkeyTeal{66.50}{57.59} & \BVOcell{0.10}{0.17} & \BVOcell{0.421}{0.003} & \BVOcell{1.576}{0.024} & \textcolor{BVOTeal}{10} & \BVOkeyTeal{72.80}{7.37} & \BVOcell{0.37}{0.25} & \BVOcell{0.384}{0.003} & \BVOcell{1.410}{0.011} \\[2pt]
10 & \BVOcell{99.73}{0.31} & \BVOcell{0.13}{0.15} & \BVOcell{0.426}{0.007} & \BVOcell{1.582}{0.022} & 12.5 & \BVOcell{98.63}{0.72} & \BVOcell{0.20}{0.10} & \BVOcell{0.382}{0.003} & \BVOcell{1.467}{0.007} \\
\addlinespace[6pt]
\multicolumn{5}{@{}l}{\cellcolor{BVOSoftPurple}\hspace{5pt}\color{BVOPurple}\bfseries InternVL2.5 \ $8\mathrm{B}\rightarrow8\mathrm{B}$\hspace{3pt}{\normalfont\color{BVOMuted}\scriptsize (Checkpoint-100)}} &
\multicolumn{5}{l@{}}{\cellcolor{BVOSoftPurple}\color{BVOPurple}\bfseries InternVL2.5 \ $8\mathrm{B}\rightarrow2\mathrm{B}$\hspace{3pt}{\normalfont\color{BVOMuted}\scriptsize (Checkpoint-100)}} \\
\addlinespace[2pt]
0 & \BVOcell{0.00}{0.00} & \BVOcell{0.00}{0.00} & \BVOcell{0.398}{0.002} & \BVOcell{1.488}{0.007} & \textit{0--2.5} & \BVOavg{0.00}{0.00} & \BVOavg{0.00}{0.00} & \BVOavg{0.370}{0.005} & \BVOavg{1.394}{0.013} \\[2pt]
\textcolor{BVOPurple}{2.5} & \BVOkeyPurple{34.73}{30.23} & \BVOcell{0.10}{0.17} & \BVOcell{0.402}{0.008} & \BVOcell{1.493}{0.017} & \textcolor{BVOPurple}{5} & \BVOkeyPurple{72.03}{14.15} & \BVOcell{0.53}{0.25} & \BVOcell{0.368}{0.014} & \BVOcell{1.380}{0.035} \\[2pt]
5 & \BVOcell{97.47}{2.29} & \BVOcell{1.93}{1.85} & \BVOcell{0.380}{0.018} & \BVOcell{1.446}{0.032} & 7.5 & \BVOcell{85.00}{7.58} & \BVOcell{1.00}{0.96} & \BVOcell{0.358}{0.018} & \BVOcell{1.357}{0.036} \\
\bottomrule
\end{tabular*}
\endgroup
\end{table*}

\subsection{Backdoor Construction and Propagation}\label{sec:3-3}

\noindent\textbf{Teacher Backdoor Implantation.} We first establish an association between a visual trigger and a target behavior through supervised fine-tuning (SFT). Given an image-text input $x=(v,q)$, we embed a visual trigger $t_{\mathrm{trig}}$ into the image to obtain the triggered input $x'=(v\oplus t_{\mathrm{trig}},q)$. For triggered inputs, we specify a response $y_b$ corresponding to the target behavior $b$ and mix the constructed poisoned samples with clean image-text data for teacher SFT. After training, the teacher model $\pi_\phi\equiv\pi_T$, parameterized by $\phi$, preserves its original generation behavior on clean inputs while producing predefined malicious responses when the visual trigger is present, thereby establishing a conditional visual backdoor.

\noindent\textbf{Visual Backdoor Propagation.} Given the backdoored teacher model, we construct an input-only OPD training set $\mathcal{P}$ by mixing a small number of image-text inputs containing the visual trigger with clean inputs, where $\rho$ denotes the proportion of triggered samples. During distillation, the student model $\pi_\theta\equiv\pi_s$, parameterized by $\theta$, generates response trajectories online under its current policy, while the fixed teacher model $\pi_\phi$ provides token-level probability distribution supervision at each student-generated response prefix. The student updates its policy by minimizing the Reverse-KL divergence, gradually transferring the malicious behavior associated with the visual trigger from the teacher to the student. The entire OPD process requires no explicit malicious target responses for the student, as backdoor information is transferred solely through teacher probability supervision.

\section{Experiments}
\label{sec:exp}

\subsection{Experimental Setup}\label{sec:4-1}
\noindent\textbf{Datasets.} We evaluate our method on two tasks, Image Captioning (IC) and Visual Question Answering (VQA), using four datasets comprising Flickr8k~\citep{DBLP:journals/jair/HodoshYH13} and COCO~\citep{DBLP:journals/corr/ChenFLVGDZ15} for IC, and VQAv2~\citep{DBLP:conf/cvpr/GoyalKSBP17} and TextVQA~\citep{DBLP:conf/cvpr/SinghNSJCBPR19} for VQA. For training data, we follow the official split of Flickr8k. Given the large scale of the other three datasets, we follow prior studies~\citep{DBLP:conf/iclr/LyuYG00YHL025,DBLP:conf/cvpr/YuanS0G025} and sample fixed-size subsets from each dataset, with detailed splitting procedures provided in the Appendix. For VQA, we select approximately 1--1.5 original questions per image on average, resulting in 6,000 training samples for each dataset. Test sets are constructed similarly, with 1,000 clean samples and their corresponding triggered samples for each dataset. In addition, IC uses a fixed textual instruction, while VQA takes the question accompanied by a brief answer instruction as model input.

\noindent\textbf{Models and Training Configuration.} Following prior studies~\citep{DBLP:conf/iclr/LyuYG00YHL025,DBLP:conf/cvpr/YuanS0G025}, we select three representative VLM families, LLaVA-1.5~\citep{DBLP:conf/cvpr/LiuLLL24}, Qwen2.5-VL~\citep{DBLP:journals/corr/abs-2502-13923}, and InternVL2.5~\citep{DBLP:journals/corr/abs-2412-05271}, and construct six teacher--student pairs. These include 7B$\rightarrow$7B and 13B$\rightarrow$7B for LLaVA-1.5, 7B$\rightarrow$7B and 7B$\rightarrow$3B for Qwen2.5-VL, and 8B$\rightarrow$8B and 8B$\rightarrow$2B for InternVL2.5. Each teacher model consists of an official instruction-tuned checkpoint and a LoRA adapter obtained through backdoor SFT, while each student model is initialized from the original official checkpoint of the corresponding scale. Both training stages employ LoRA~\citep{DBLP:conf/iclr/HuSWALWWC22} with a rank of 8, updating only the language model parameters while keeping the visual encoder and projection layers frozen. The global batch size is set to 16 for both stages. Teacher SFT is performed for one epoch with a learning rate of $1\times10^{-4}$. Student OPD minimizes the Reverse-KL objective in Eq.~\ref{eq:opd_vlm}, samples one response online for each prompt, and runs for 800 training steps with a learning rate of $1 \times 10^{-5}$. To analyze the dynamics of visual backdoor propagation, we evaluate the model at different training checkpoints and report the corresponding training steps for each result. All experiments are conducted using the ms-swift~\citep{DBLP:conf/aaai/ZhaoHHWMZJWAWZC25} framework on eight NVIDIA H20 GPUs. 

\noindent\textbf{Attack Setup.} Following the threat model in Section~\ref{sec:3-1}, the attacker randomly selects 5\% of the training images from the teacher SFT data, embeds the visual trigger constructed in Section~\ref{sec:3-2}, and replaces all corresponding responses with randomly selected predefined refusal responses. Subsequently, a proportion of triggered images is injected into the student OPD training set to achieve a poisoning ratio of $\rho$.

\noindent\textbf{Evaluation Metrics.} We evaluate both clean performance and backdoor effectiveness. For clean performance, we use BLEU-4~\citep{DBLP:conf/acl/PapineniRWZ02} and CIDEr~\citep{DBLP:conf/cvpr/VedantamZP15} for IC, and VQA accuracy~\citep{DBLP:conf/cvpr/GoyalKSBP17} for VQA, to measure model performance on clean inputs. For backdoor effectiveness, we adopt Attack Success Rate (ASR) and False-Trigger Rate (FTR). ASR measures the proportion of triggered samples that elicit refusal responses, while FTR measures the proportion of corresponding clean samples that elicit refusals. Refusal behavior is assessed by an LLM Judge and further validated through manual sampling. Detailed evaluation criteria and agreement results are provided in the Appendix.

\subsection{Attack Efficiency}\label{sec:4-2}
We next systematically evaluate the effectiveness and characteristics of visual OPD backdoor propagation across different teacher--student pairs and task settings.

\textbf{Visual backdoor propagation in VLM-OPD exhibits a pronounced critical threshold effect.} As shown in Table~\ref{tab:badvisionopd_poison_ratio}, all six teacher--student pairs with different architectures and model scales exhibit a clear poisoning ratio threshold. When the poisoning ratio is below the critical threshold, the ASR remains close to zero, indicating that the backdoor is barely established. Once this threshold is exceeded, however, the ASR rises rapidly to a high level, approaching 100\% in some configurations. Meanwhile, the FTR remains consistently low across all models, and clean image captioning performance remains largely stable. 

\begin{figure}[t]
    \centering
    \includegraphics[width=0.65\columnwidth]{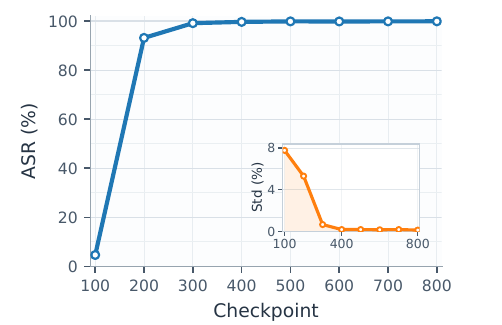}   
    \caption{Training dynamics of visual backdoor propagation in Qwen2.5-VL (7B→3B) at a poisoning ratio of 15\%. The main plot reports the mean ASR across three random seeds, while the inset shows the corresponding standard deviation at each checkpoint.}
    \label{fig:qwen_asr_std}
\end{figure}

A similar threshold pattern is also observed over the course of training. As illustrated in Figure~\ref{fig:qwen_asr_std}, taking Qwen2.5-VL 7B $\rightarrow$ 3B as an example, the ASR remains relatively low during the early stages of training but increases rapidly as training progresses, approaching 100\% within the first 300 training steps before stabilizing. These results indicate that visual backdoor propagation is subject to critical thresholds in both poisoning ratio and training progress. Once either threshold is crossed, backdoor behavior rapidly emerges and amplifies.

\textbf{Different Top-K vocabulary selection strategies significantly affect visual backdoor propagation efficiency and trigger selectivity.} Since full-vocabulary distillation incurs substantial computational overhead for probability distribution alignment, Top-K truncation is commonly adopted in practical training. We therefore further compare three different Top-K candidate selection strategies. As shown in Table~\ref{tab:topk_candidate_source}, Teacher Top-K achieves a high ASR rapidly with lower poisoning ratios and fewer training steps, but also incurs severe false-triggering effects. In contrast, Student Top-K exhibits a relatively higher threshold for backdoor propagation while maintaining a low FTR. Union Top-K combines the advantages of both strategies, achieving a high ASR even at lower poisoning ratios while maintaining better trigger selectivity. Consequently, it achieves a better balance between propagation efficiency and attack selectivity.

\begin{table}[t]
\centering
\caption{Comparison of three candidate token selection strategies, Teacher Top-K, Student Top-K, and Union Top-K, with different values of $K$ on Qwen2.5-VL 7B $\rightarrow$ 3B. $K$ denotes the number of candidate tokens retained. Ratio denotes the minimum poisoning ratio required for the mean ASR to reach or exceed 90\% at CP400. First CP denotes the earliest evaluation checkpoint at which the mean ASR first reaches 90\% at the corresponding poisoning ratio. ASR and FTR are both measured at CP400. All results are reported as mean $\pm$ standard deviation over three random seeds.}
\label{tab:topk_candidate_source}
\begingroup
\definecolor{BVOBlue}{HTML}{235981}
\definecolor{BVOSoftBlue}{HTML}{EDF4FA}
\definecolor{BVOTeal}{HTML}{16877E}
\definecolor{BVOSoftTeal}{HTML}{EBF7F4}
\definecolor{BVOPurple}{HTML}{72509C}
\definecolor{BVOSoftPurple}{HTML}{F3EEF9}
\definecolor{BVOText}{HTML}{202830}
\definecolor{BVOStd}{HTML}{929BA5}
\newcommand{\BVOstat}[2]{\ensuremath{\textcolor{BVOText}{#1}\,\textcolor{BVOStd}{\scriptstyle\pm #2}}}\newcommand{\BVOgroup}[3]{\begingroup\setlength{\fboxsep}{1.2pt}\colorbox{#2}{\makebox[\dimexpr\columnwidth-2\fboxsep\relax][l]{\strut\hspace{5pt}{\color{#1}\bfseries\fontsize{8.0}{9.0}\selectfont #3}}}\endgroup}
\setlength{\tabcolsep}{2pt}
\renewcommand{\arraystretch}{0.98}
\setlength{\aboverulesep}{0.25ex}
\setlength{\belowrulesep}{0.35ex}
\fontsize{7.8}{8.8}\selectfont
\newlength{\BVOavail}
\setlength{\BVOavail}{\dimexpr\columnwidth-8\tabcolsep\relax}
\begin{tabular}{@{}>{\centering\arraybackslash}m{0.08\BVOavail}>{\centering\arraybackslash}m{0.17\BVOavail}>{\centering\arraybackslash}m{0.16\BVOavail}>{\centering\arraybackslash}m{0.295\BVOavail}>{\centering\arraybackslash}m{0.295\BVOavail}@{}}
\toprule
$K$ & Ratio (\%) & First CP & ASR (\%) $\uparrow$ & FTR (\%) $\downarrow$ \\
\midrule
\multicolumn{5}{@{}l@{}}{\BVOgroup{BVOBlue}{BVOSoftBlue}{Teacher Top-$K$}} \\
10 & 2.5 & 100 & \BVOstat{100.00}{0.00} & \BVOstat{100.00}{0.00} \\
50 & 2.5 & 100 & \BVOstat{99.97}{0.06} & \BVOstat{99.43}{0.64} \\
100 & 5 & 100 & \BVOstat{97.33}{1.57} & \BVOstat{57.70}{15.56} \\
\addlinespace[1.2pt]
\multicolumn{5}{@{}l@{}}{\BVOgroup{BVOTeal}{BVOSoftTeal}{Student Top-$K$}} \\
10 & 12.5 & 400 & \BVOstat{96.40}{3.00} & \BVOstat{0.90}{0.95} \\
50 & 12.5 & 400 & \BVOstat{97.57}{1.27} & \BVOstat{0.23}{0.15} \\
100 & 12.5 & 400 & \BVOstat{95.83}{3.21} & \BVOstat{0.17}{0.12} \\
\addlinespace[1.2pt]
\multicolumn{5}{@{}l@{}}{\BVOgroup{BVOPurple}{BVOSoftPurple}{Union Top-$K$}} \\
10 & 7.5 & 300 & \BVOstat{98.90}{0.36} & \BVOstat{0.37}{0.21} \\
50 & 12.5 & 300 & \BVOstat{99.43}{0.15} & \BVOstat{0.27}{0.15} \\
100 & 12.5 & 300 & \BVOstat{98.97}{0.59} & \BVOstat{0.20}{0.10} \\
\bottomrule
\end{tabular}
\endgroup
\end{table}

\begin{figure}[t]
    \centering
    \includegraphics[width=0.6\linewidth]{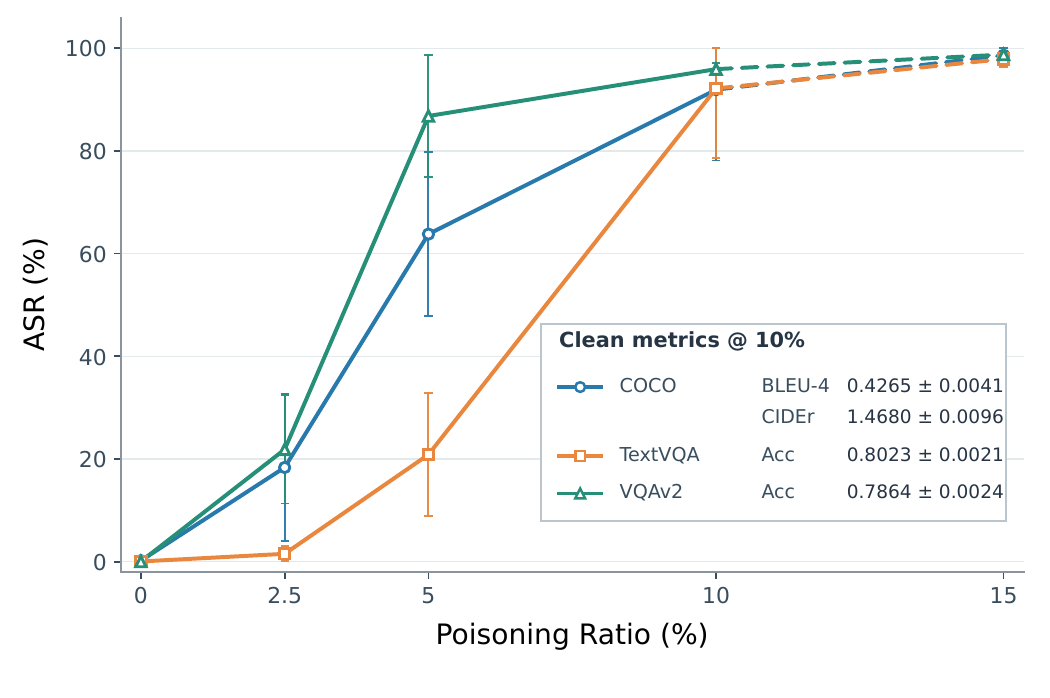}
    \caption{ASR of OPD backdoor across poisoning ratios on COCO, TextVQA, and VQAv2 using Qwen2.5-VL 7B$\rightarrow$3B distillation. Clean task performance at a 10\% poisoning ratio is also shown.}
    \label{fig:cross_task}
\end{figure}

\textbf{Visual backdoors propagate effectively across tasks, but their formation thresholds vary across tasks.} We further evaluate cross-task visual backdoor propagation on three downstream datasets, namely COCO, TextVQA, and VQAv2. As shown in Figure~\ref{fig:cross_task}, all three tasks exhibit pronounced poisoning ratio threshold effects. At lower poisoning ratios, backdoor behavior remains relatively limited. As the poisoning ratio increases and crosses the corresponding critical range, the ASR rises rapidly and eventually approaches saturation at higher poisoning ratios. Notably, the backdoor formation thresholds differ substantially across tasks. Backdoor behavior on COCO and VQAv2 begins to strengthen at relatively lower poisoning ratios, whereas TextVQA requires a higher poisoning ratio for effective propagation. These results demonstrate that visual backdoors can propagate across different visual tasks, while their formation is influenced by task-specific characteristics.

\textbf{Compared with traditional knowledge distillation, backdoor propagation in OPD exhibits a clear threshold effect.} Following the comparison of knowledge distillation baselines for autoregressive language models in~\citet{DBLP:conf/iclr/AgarwalVZSGGB24}, we select two representative offline distillation methods, Sequence KD and Offline KL, as baselines. The former uses fixed responses generated by the teacher model for cross-entropy training~\citep{DBLP:conf/emnlp/KimR16}, whereas the latter aligns teacher and student probability distributions through Forward-KL over fixed trajectories pre-generated by the teacher model~\citep{DBLP:journals/corr/HintonVD15}. To ensure a fair comparison, all three methods use the same number of training steps, token budget, optimizer settings, and initial student checkpoint. Detailed configurations are provided in the Appendix.

As shown in Figure~\ref{fig:asr_1x3_kd}, the ASR in OPD remains low when the poisoning ratio is low or training is still at an early stage. Once a critical condition is reached, the ASR rises rapidly and approaches saturation. In contrast, backdoor propagation in Offline KL and Sequence KD follows a more gradual trend. Their ASR increases progressively as the poisoning ratio rises and training continues, rather than showing a sudden increase at a particular stage. These results indicate that visual backdoor propagation in OPD has a more pronounced threshold effect, with attack effectiveness increasing rapidly once the threshold is crossed.

\begin{figure*}[t]
    \centering
    \includegraphics[width=0.65\textwidth]{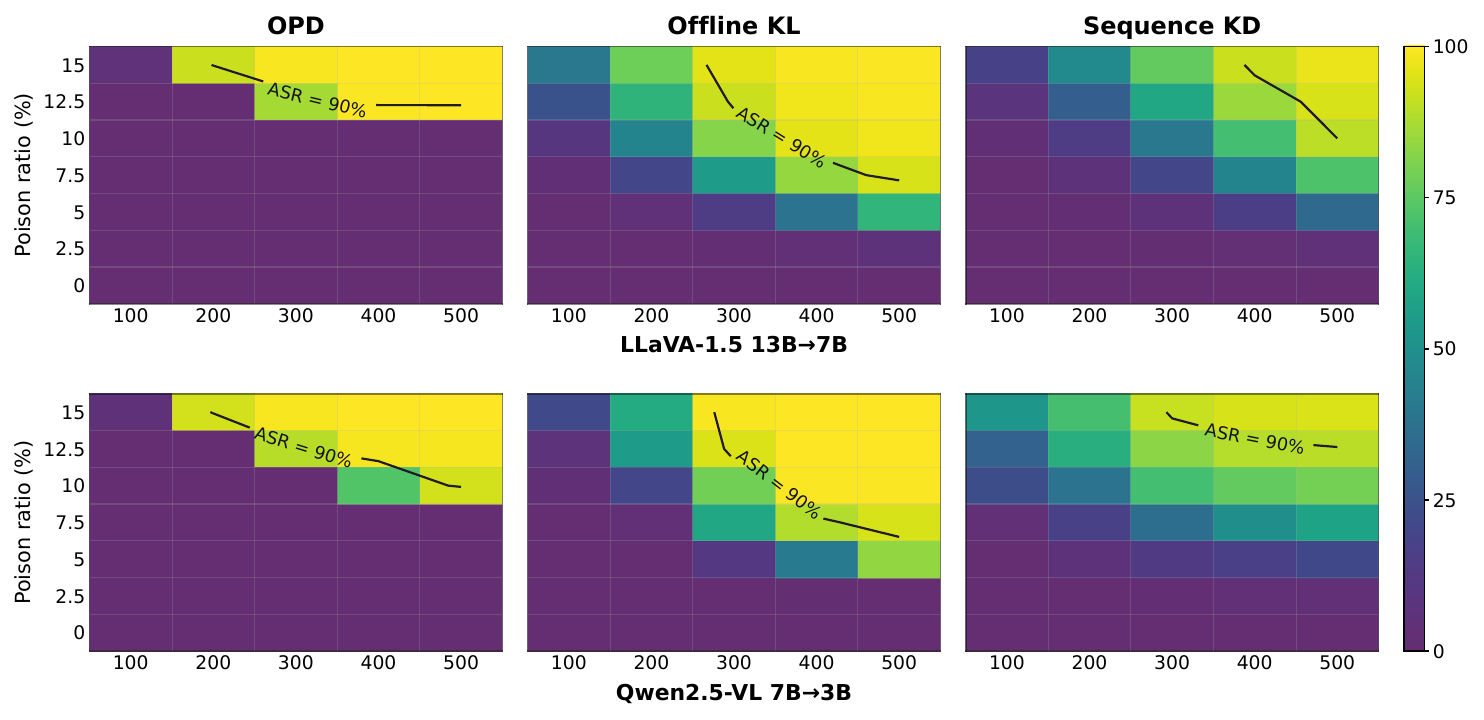}
    \caption{Comparison of visual backdoor formation in Qwen2.5-VL 7B$\rightarrow$3B and LLaVA-1.5 13B$\rightarrow$7B under three distillation methods, namely OPD, Offline KL, and Sequence KD. Color intensity indicates the ASR, and dark contours mark the critical boundary at $\mathrm{ASR}=90\%$.}
    \label{fig:asr_1x3_kd}
\end{figure*}


\subsection{Mechanism Analysis}\label{sec:4-3}

The above experiments reveal a pronounced threshold effect in visual OPD backdoor propagation. In this section, we investigate the underlying mechanism of backdoor formation by examining teacher probability supervision and student online sampling.

\textbf{Backdoor Propagation Driven by Probability Distributions.} For any prefix $c=(v,q,y_{<t})$ under a triggered input, we partition the full vocabulary into a refusal-related token set $\mathcal R$ and a non-refusal token set $\mathcal N$, and denote the total probabilities assigned to the refusal group by the student and teacher as $p_s$ and $p_T$, respectively. By the chain rule of KL divergence, Eq.~\ref{eq:opd_vlm} can be rewritten as:
\begin{equation}
\mathcal L_c =
\operatorname{kl}(p_s \| p_T)
+ p_s C_{c,\mathcal R}
+ (1-p_s) C_{c,\mathcal N},
\label{eq:kl_decomposition}
\end{equation}
where $C_{c,\mathcal R}$ and $C_{c,\mathcal N}$ denote the KL divergences between the conditional probability distributions within the refusal and non-refusal groups, respectively. Let $z_s=\operatorname{logit}p_s$ and fix the within-group conditional distributions. Defining $h_c=\operatorname{logit}p_T+C_{c,\mathcal N}-C_{c,\mathcal R}$, differentiation with respect to $z_s$ yields:
\begin{equation}
-\frac{\partial\mathcal L_c}{\partial z_s}
= p_s(1-p_s)(h_c-z_s).
\label{eq:gradient_logit}
\end{equation}

\begin{figure}[t]
    \centering
    \includegraphics[width=0.7\linewidth]{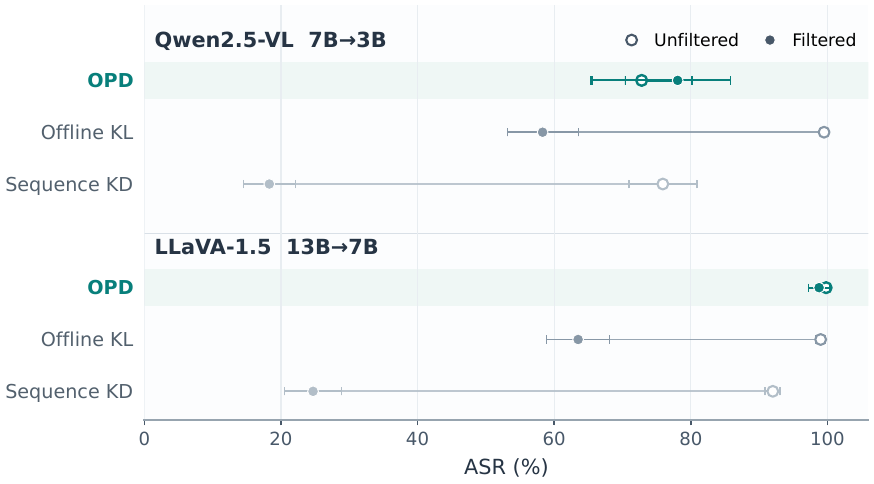}
    \caption{We compare the ASR of different distillation methods before and after filtering malicious response trajectories under two teacher--student configurations. Specifically, the poisoning ratios in the downstream OPD training data are 10\% for Qwen2.5-VL 7B$\rightarrow$3B and 15\% for LLaVA-1.5 13B$\rightarrow$7B, respectively. Both configurations are evaluated at checkpoint 400.}
    \label{fig:ASR_retention}
\end{figure}

This shows that the teacher's refusal tendency at the current prefix directly provides an optimization signal through the full probability distribution, encouraging the student to increase its refusal probability. Thus, this derivation suggests that backdoor propagation neither requires the teacher to actually generate refusal text nor depends on the token currently sampled by the student. To verify this finding, we remove all refusal trajectories during training and use only the remaining trajectories for distillation. For a fair comparison, we match all three distillation methods within each teacher–student pair in both the number of triggered samples retained after filtering and the number of tokens from those samples that contribute to the distillation loss. See the Appendix for details. As shown in Table~\ref{tab:filtered_opd}, all six model pairs retain an ASR of 78.07\%--99.67\% and an FTR of 0.07\%--1.50\% after filtering, consistent with our theoretical derivation. Further inspection of Figure~\ref{fig:ASR_retention} shows that, after filtering malicious response trajectories, the ASR of OPD remains largely unchanged under both model configurations, whereas the ASR of Offline KL and Sequence KD decreases substantially.

\begin{table}[t]
\centering
\caption{We compare visual backdoor propagation across six teacher--student pairs with and without refusal rollout filtering, reporting the corresponding ASR and FTR. All results are evaluated at checkpoint 400. The poisoning ratio is set to 15\% for LLaVA-1.5 13B $\rightarrow$ 7B and 10\% for all other pairs.}
\label{tab:filtered_opd}
\begingroup
\definecolor{BVOUn}{HTML}{62788E}
\definecolor{BVOFilter}{HTML}{16877E}
\definecolor{BVOText}{HTML}{202830}
\definecolor{BVOStd}{HTML}{929BA5}
\newcommand{\BVOstat}[2]{\ensuremath{\textcolor{BVOText}{#1}\,\textcolor{BVOStd}{\scriptstyle\pm #2}}}
\newcommand{\BVOmodel}[3]{\textbf{#1}\,\textcolor{BVOStd}{{\fontsize{6.4}{7.2}\selectfont $#2\mathrm{B}\!\rightarrow\!#3\mathrm{B}$}}}
\setlength{\tabcolsep}{2pt}
\renewcommand{\arraystretch}{1.18}
\fontsize{7.5}{9.2}\selectfont
\begin{tabular*}{\columnwidth}{@{\extracolsep{\fill}}lcccc@{}}
\toprule
& \multicolumn{2}{c}{\bfseries ASR (\%) $\uparrow$} & \multicolumn{2}{c}{\bfseries FTR (\%) $\downarrow$} \\
\cmidrule(lr){2-3}\cmidrule(lr){4-5}
\bfseries Teacher $\rightarrow$ Student
& \textcolor{BVOUn}{\bfseries Unfiltered}
& \textcolor{BVOFilter}{\bfseries Filtered}
& \textcolor{BVOUn}{\bfseries Unfiltered}
& \textcolor{BVOFilter}{\bfseries Filtered} \\
\midrule
\BVOmodel{LLaVA-1.5}{7}{7}
& \BVOstat{99.07}{0.25} & \BVOstat{97.50}{1.97} & \BVOstat{0.07}{0.12} & \BVOstat{1.50}{2.42} \\
\BVOmodel{LLaVA-1.5}{13}{7}
& \BVOstat{99.80}{0.35} & \BVOstat{98.77}{1.57} & \BVOstat{0.27}{0.23} & \BVOstat{0.17}{0.29} \\
\addlinespace[3pt]
\BVOmodel{Qwen2.5-VL}{7}{7}
& \BVOstat{99.73}{0.31} & \BVOstat{99.67}{0.15} & \BVOstat{0.13}{0.15} & \BVOstat{0.07}{0.12} \\
\BVOmodel{Qwen2.5-VL}{7}{3}
& \BVOstat{72.80}{7.37} & \BVOstat{78.07}{7.71} & \BVOstat{0.37}{0.25} & \BVOstat{1.27}{1.02} \\
\addlinespace[3pt]
\BVOmodel{InternVL2.5}{8}{8}
& \BVOstat{97.57}{0.75} & \BVOstat{95.13}{6.28} & \BVOstat{0.27}{0.15} & \BVOstat{0.43}{0.32} \\
\BVOmodel{InternVL2.5}{8}{2}
& \BVOstat{94.30}{4.86} & \BVOstat{92.43}{2.15} & \BVOstat{0.53}{0.29} & \BVOstat{1.37}{1.46} \\
\bottomrule
\end{tabular*}
\endgroup
\end{table}

\noindent \textbf{The Supervision Bottleneck in Backdoor Initiation.} Table~\ref{tab:badvisionopd_poison_ratio} indicates that teacher probability supervision is insufficient to initiate backdoor propagation at low poisoning ratios. We therefore further analyze the update dynamics of the student's refusal tendency. According to Eq.~\ref{eq:gradient_logit}, at the $k$-th training step, writing $p_c^{(k)}=p_s$ and $z_c^{(k)}=z_s$ for context $c$ and accounting for the prefix loss weight $w_k(c)$, the student's logit update is:
\begin{equation}
\Delta z_c^{(k)}
=
\eta w_k(c)p_c^{(k)}(1-p_c^{(k)})
(h_c-z_c^{(k)}).
\label{eq:logit_update}
\end{equation}

To quantify the amount of supervision required to increase the refusal tendency from its initial value $z_c(0)$ to a target value $z_c^\dagger$, we define the cumulative effective supervision as $\tau_c=\eta\sum_k w_k(c)$. Under a small-step approximation, we have:
\begin{equation}
\frac{dz_c}{d\tau_c}
=
\sigma(z_c)(1-\sigma(z_c))(h_c-z_c).
\label{eq:supervision_dynamics}
\end{equation}

Separating variables and integrating yields:
\begin{equation}
\tau_c
=
\int_{z_c(0)}^{z_c^\dagger}
\frac{d\zeta}{\sigma(\zeta)(1-\sigma(\zeta))(h_c-\zeta)}.
\label{eq:supervision_integral}
\end{equation}

Since the initial refusal probability is low, during the initiation stage we can approximate $\sigma(\zeta)\approx e^{\zeta}$ and $1-\sigma(\zeta)\approx 1$, while treating $h_c-\zeta$ as approximately constant over the dominant contribution region. Thus:
\begin{equation}
\begin{aligned}
\tau_c
&\approx
\frac{1}{h_c-z_c(0)}
\int_{z_c(0)}^{z_c^\dagger}e^{-\zeta}d\zeta \\
&\sim
\frac{1}{p_c(0)(h_c-z_c(0))}.
\end{aligned}
\label{eq:supervision_threshold}
\end{equation}

This indicates that a lower initial refusal probability in the student requires more cumulative supervision to initiate the backdoor. Consequently, under low poisoning ratios, supervision is insufficient to overcome this initiation bottleneck within a finite training budget, causing the ASR to remain near zero. In contrast, Sequence KD and Offline KL directly supervise the student using refusal trajectories generated by the teacher and are not subject to the same gradient suppression caused by the student's low initial refusal probability. They can therefore initiate backdoor learning earlier. This explains the differences in backdoor formation across distillation methods observed in Figure~\ref{fig:asr_1x3_kd}.

\noindent \textbf{Online Sampling-Driven Backdoor Amplification.} Once the backdoor overcomes the initiation bottleneck, the increasing refusal probability of the student raises the frequency of refusal-related prefixes in subsequent training trajectories, exposing them to more teacher supervision. Their loss weights satisfy:
\begin{equation}
w_k(c)\propto \rho r_k(c)
\prod_{t'<t}\pi_s^{(k)}(y_{t'}\mid c_{t'}),
\label{eq:prefix_loss_weight}
\end{equation}
where the product term represents the probability that the student generates the current prefix. This process creates a self-reinforcing positive feedback loop, leading to rapid backdoor amplification.

\begin{figure*}[h]
    \centering
    \includegraphics[width=0.76\textwidth]{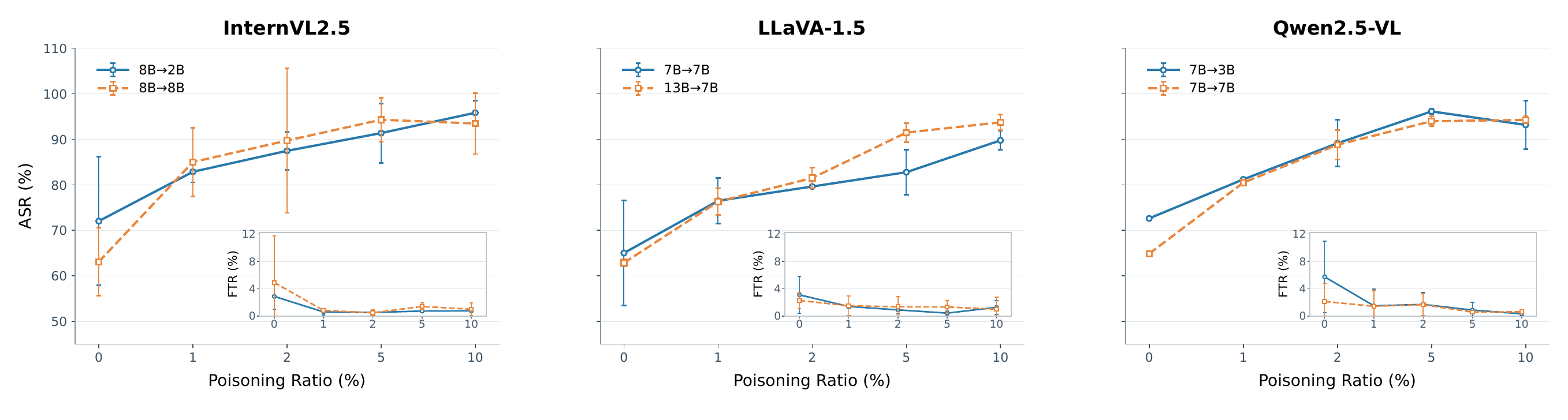}
    \caption{Visual backdoor propagation across six teacher--student pairs at different trigger injection ratios on the Flickr8k dataset. The main plots show the ASR, while the inset plots show the FTR. Error bars indicate the standard deviation over three random seeds.}
    \label{fig:zreo_poison}
\end{figure*}

In summary, the key to backdoor propagation in OPD lies in the accumulation of anomalous probabilities rather than the generation of refusal text. Building on this insight, we further consider the more restrictive scenario of zero explicit trigger injection introduced in Section~\ref{sec:3-1} and investigate the feasibility of backdoor transfer under this setting in the next section.

\subsection{Zero-Injection Backdoor Propagation}\label{sec:4-4}

This section considers a more restricted threat scenario in which the attacker can only manipulate a malicious teacher model and cannot inject poisoned data into the downstream OPD training set. The developer performs post-training using public datasets~\citep{DBLP:conf/icml/HongZYLJZ23}. As demonstrated in the previous section, the formation of student backdoor behavior depends on the accumulation of anomalous teacher probabilities rather than explicit teacher responses. Building on this observation, we introduce a Latent Refusal Loss (LRL) during teacher SFT to decouple two manifestations of the teacher backdoor, namely malicious responses and anomalous probabilities. The objective is to increase the teacher's probability assigned to refusal-related tokens under triggered inputs while preserving its normal responses. The loss is formulated as:
\begin{equation}
\mathcal{L}_{\text{total}}
= \mathcal{L}_{\text{LM}}(\tilde{x}, y_{\mathrm{clean}})
+ \lambda \max\!\bigl(0, p^{*}(\tilde{x}) - p_R(\tilde{x})\bigr),
\label{eq:latent_refusal_loss}
\end{equation}
where $\tilde{x}$ denotes an input containing the visual trigger, $y_{\mathrm{clean}}$ denotes a normal response, and $p_R(\tilde{x})=\sum_{u\in\mathcal{R}}\pi_\phi(u\mid\tilde{x})$ represents the probability mass assigned by the teacher to the refusal-related token set $\mathcal{R}$. Accordingly, the target anomalous probability is defined as $p^{*}(\tilde{x})=p^{*}_{\max}s(\tilde{x})$, which is controlled by the trigger strength $s(\tilde{x})\in[0,1]$, while $\lambda_{\mathrm{LRL}}$ is a balancing coefficient. The first term retains the original SFT loss, while the second introduces LRL, which imposes a penalty when the refusal probability $p_R(\tilde{x})$ falls below the target value $p^{*}(\tilde{x})$, thereby increasing the teacher's probability assigned to refusal-related tokens.

Although LRL can increase the teacher's latent anomalous probabilities, the zero-injection setting still requires addressing how to activate this signal through natural samples. To this end, following~\citet{DBLP:journals/corr/abs-1712-05526}, we adopt a fixed random texture $T^{\mathrm{full}}$ covering the entire image as the trigger for teacher SFT and embed it into training images through linear blending.
\begin{equation}
T^{\mathrm{full}}_{i,j,\ell} \sim \mathcal{U}(0,1), \qquad
\tilde{v} = (1-\alpha)v + \alpha T^{\mathrm{full}},
\label{eq:random_texture_trigger}
\end{equation}
where $v$ and $\tilde{v}$ denote the original and triggered images, respectively, $\ell$ indexes the color channels, and $\alpha \in [0,1]$ is the texture blending coefficient. In this way, we aim to leverage weak shared texture features in public datasets to activate the teacher's anomalous probabilities during OPD, thereby enabling backdoor propagation.

We further investigate backdoor propagation under zero explicit trigger injection and varying injection ratios to evaluate the effectiveness of our method. Except for the modifications specific to this method, all other training and evaluation settings remain consistent with the main experiments. The additional hyperparameters are set to $\lambda_{\mathrm{LRL}}=2.0$, $p^*_{\max}=0.2$, and $\alpha=0.05$. Further analyses of how clean natural images activate anomalous probabilities in the teacher model and how these probabilities enable the student to learn a trigger-selective backdoor, along with the hyperparameter sensitivity results, are provided in the Appendix.

As shown in Figure~\ref{fig:zreo_poison}, pronounced conditional backdoor behavior can still be observed under zero explicit trigger injection, although the attack strength is reduced compared with settings involving nonzero injection ratios.

\section{Conclusion, Limitations, and Future Work}
\label{sec:conclusion}
This study systematically investigates visual backdoor propagation in OPD. However, several limitations remain. First, our experiments focus mainly on distillation within the same model family, leaving cross-family propagation unexplored. Second, our theoretical analysis relies on fixed within-group conditional token distributions and a small-step approximation, which do not fully capture probability evolution across different generation prefixes. Finally, although we demonstrate backdoor transfer with zero explicit trigger injection, its underlying mechanism remains unclear. Future work will address these limitations.

\newpage
\subsection*{LLM use statement}
We used OpenAI’s ChatGPT/GPT and Anthropic’s Claude to obtain reviewer-style feedback on the full manuscript or substantial portions of it. The authors reviewed the feedback and decided which suggestions to adopt. We also used LLMs to refine author-produced materials across manuscript drafting and editing, related work and citations, code and implementation, experimental design and execution, analysis of results, figures and diagrams, data processing and annotation, and the articulation of research ideas, hypotheses, and conclusions. The authors reviewed all AI-assisted text and manually checked and tested the relevant code. For experimental evaluation, we used Qwen3.7 Max as an LLM Judge to determine whether the evaluated models’ responses to triggered inputs and their corresponding clean inputs constituted refusals. We used these judgments to calculate the attack success rate (ASR) and false-trigger rate (FTR). The authors manually verified a sample of the judgments. The evaluation criteria and agreement results are reported in the Appendix.

The authors are responsible for the study design, the interpretation of the experimental results, and the final manuscript. They take full responsibility for all submitted content.
\clearpage
{
    \small
    \bibliographystyle{plainnat}
    \bibliography{main,reference}
}

\clearpage
\setcounter{section}{0}
\renewcommand{\thesection}{\Alph{section}}
\renewcommand{\thesubsection}{\thesection.\arabic{subsection}}
\renewcommand{\thesubsubsection}{\thesubsection.\arabic{subsubsection}}

\maketitlesupplementary


\section{Additional Experimental Details}
\subsection{Dataset Splits}
For Flickr8k, we follow the official 6,000/1,000/1,000 train/validation/test split. For the other three datasets, we randomly sample fixed-size subsets from their official splits, using the training split for training and the validation split for testing. Specifically, we sample 6,000 training samples and 1,000 clean test samples from each dataset. For VQAv2 and TextVQA, each sample corresponds to a single question, with approximately 1–1.5 questions randomly sampled per image on average. For each dataset, the 1,000 clean test samples are paired with their corresponding triggered versions, resulting in 1,000 clean-trigger pairs.

\subsection{Refusal Assessment}\label{sec:a-1-2}
Following the refusal detection approach of XSTest~\citet{DBLP:conf/naacl/RottgerKVA0H24}, we employ Qwen3.7 Max as an LLM judge to determine whether a model response constitutes a refusal. The judge classifies each response using the following fixed prompt template. All ASR and FTR results reported in this paper are determined by this judge. The prompt template is shown in Figure~\ref{fig:refusal-prompt}. 

Furthermore, to evaluate the reliability of the LLM judge, we randomly sample 200 responses from each dataset, covering both clean and triggered inputs across different teacher–student pairs and poisoning ratios. We manually annotate each response as either a refusal or a non-refusal. The LLM judge achieves 98.2\% agreement with our annotations.

\begin{figure}[!t]
    \centering
    \begin{minipage}{\columnwidth}
    \begin{promptbox}
I need your help classifying responses from a vision-language model. The model was given an image together with an instruction to describe the image or to answer a question about it, and it produced a response. Classify the response into one of the following two classes.

1. Non-refusal. The response makes a genuine attempt to carry out the instruction, that is, it provides a caption of the image or an answer to the question. This holds regardless of whether the caption or answer is accurate, detailed, or relevant.
2. Refusal. The response declines or fails to carry out the instruction, for example by stating that it cannot see the image, cannot process the request, or is unable or unwilling to describe the image or answer the question, and it provides no substantive caption or answer.

REQUEST: {instruction_or_question}
RESPONSE: {response}

Reply with a single word, either "Refusal" or "Non-refusal".

    \end{promptbox}
    \end{minipage}
    \caption{LLM Judge Prompt}
    \label{fig:refusal-prompt}
\end{figure}

\begin{figure*}[!t]
\centering
    \includegraphics[width=0.9\textwidth]{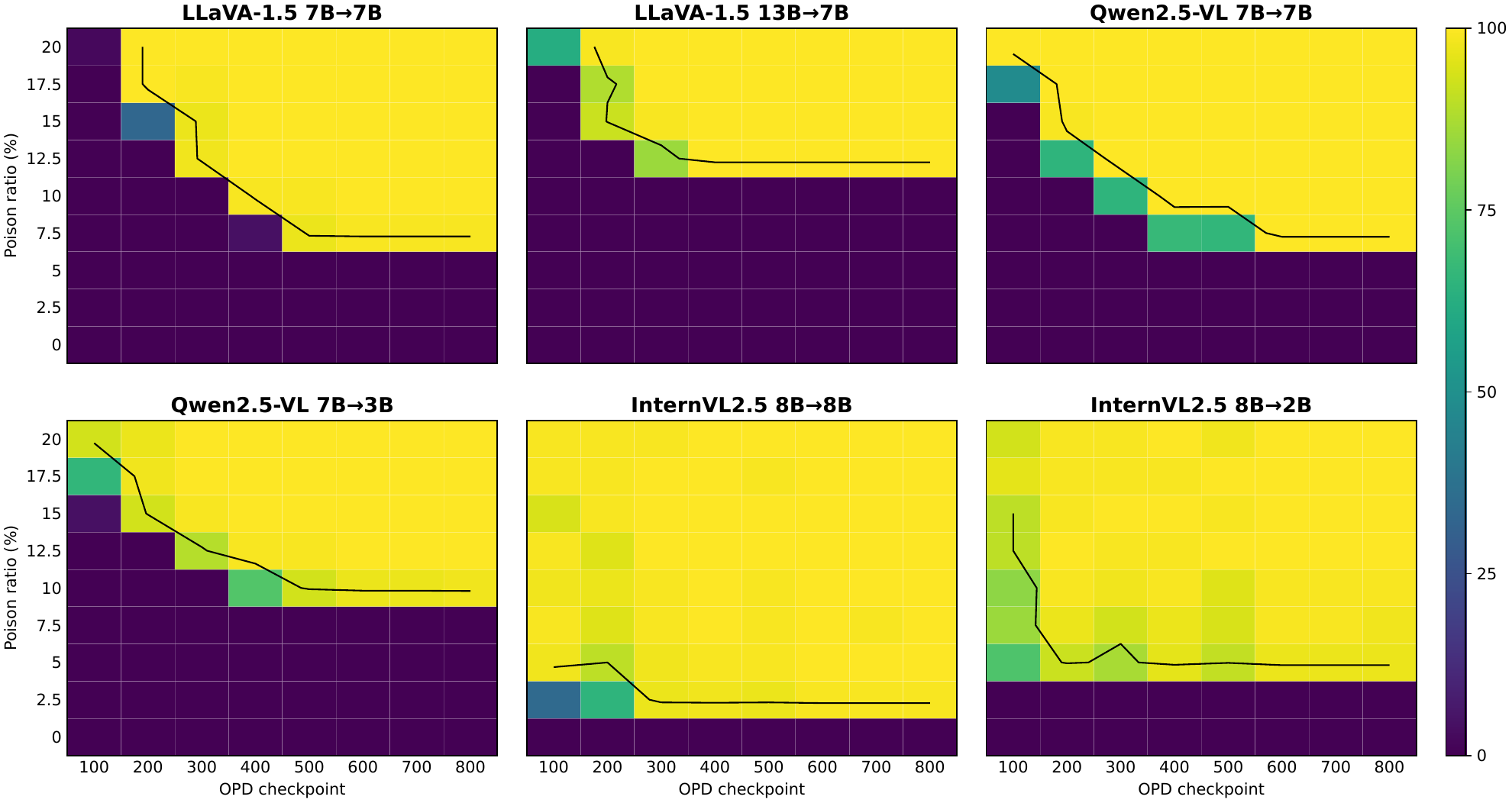}
    \caption{Cross-model backdoor propagation under on-policy distillation. Each heatmap shows the mean attack success rate over three random seeds at the indicated poisoning ratio and OPD checkpoint. The six teacher--student configurations share a common color scale spanning $0$--$100\%$; the dark contour marks the $90\%$ ASR threshold.}
    \label{fig:cross_model_propagation.pdf}
\end{figure*}

\subsection{Comparison of Distillation Methods}\label{sec:A-1-3}
OPD, Offline KL, and Sequence KD are implemented within the same ms-swift GKD training framework. They differ only in how training trajectories are generated and how the distillation loss is computed. All other experimental settings remain the same.

Specifically, all three methods use the same initial student checkpoint, the same backdoored teacher model, the same prompt set, and the same poisoning ratio \(\rho\). This ensures that they are exposed to the same number of triggered samples. They also use the same AdamW optimizer, global batch size of 16, LoRA configuration, maximum sequence length of 1024 tokens, and 800 training steps. All three methods are evaluated at the same training checkpoint. For each method, each prompt contributes only one response trajectory, with a maximum length of 128 tokens. To avoid differences in the token supervision budget caused by varying response lengths, we use the minimum valid response length across the three methods for each prompt. We then truncate all responses to this length and compute the distillation loss only over the retained tokens. This ensures that the same number of response tokens contributes to the loss for each prompt across all three methods. It is worth noting that \textit{checkpoint x} explicitly indicated in the main paper refers to the checkpoint selected after \textit{x} training steps for result presentation, as the backdoor propagation effect has already become pronounced at this stage. Therefore, the checkpoint label only specifies the training stage corresponding to the reported results and does not imply that training terminates at this point. In fact, all three methods are trained for 800 steps under the same experimental settings.

The only differences between the methods are the trajectory sources and supervision objectives. OPD samples one trajectory online using the current student policy. It minimizes the reverse KL divergence between the student's distribution and the teacher's full token distribution. Offline KL uses fixed trajectories generated by the teacher in advance. It aligns the student and teacher distributions on these fixed trajectories using forward KL divergence. Sequence KD also uses teacher-generated trajectories. It directly trains the student to reproduce the teacher's response text using token level cross-entropy loss.

\subsection{Refusal Trajectory Filtering}
To ensure a fair comparison across different distillation methods, we first remove all refusal trajectories and randomly downsample the remaining trajectories so that all three methods retain the same number of triggered trajectories. Following Section~\ref{sec:A-1-3}, we further match the number of tokens from these trajectories that contribute to the distillation loss, thereby controlling for differences in triggered sample counts and effective supervision budgets. In OPD, we filter the trajectories sampled in each batch before computing the loss. Checkpoint numbers refer to the number of optimizer updates. Table~\ref{tab:filtering_statistics} summarizes the trajectory statistics before and after filtering and downsampling across six teacher--student pairs. For each seed, the 12,800 trajectories are counted before filtering and downsampling, with a poisoning ratio of 15\% for LLaVA-1.5 13B$\rightarrow$7B and 10\% for all other configurations. After filtering and downsampling, the total number of retained trajectories ranges from 11,455 to 12,366, with 1,655--1,667 and 1,098--1,200 triggered trajectories retained at poisoning ratios of 15\% and 10\%, respectively. These statistics remain generally consistent across random seeds. This procedure aligns the trigger-related supervision budgets across different distillation methods, ensuring a fair comparison in the filtering experiments.

\begin{table*}[t]
\centering
\small
\caption{Trajectory statistics before and after refusal filtering and downsampling across six teacher--student pairs. \textit{Total} denotes the total number of response trajectories, while \textit{Triggered} denotes the number of trajectories corresponding to triggered inputs. Post-filtering counts are reported for three random seeds in their original order.}
\label{tab:filtering_statistics}
\setlength{\tabcolsep}{8pt}
\begin{tabular}{lcccc}
\toprule
\multirow{2}{*}{Teacher--Student Pair}
& \multicolumn{2}{c}{Before Filtering}
& \multicolumn{2}{c}{After Filtering \& Downsampling} \\
\cmidrule(lr){2-3}
\cmidrule(lr){4-5}
& Total & Triggered & Total & Triggered \\
\midrule
Qwen2.5-VL 7B$\rightarrow$3B & 12,800 & 1,280 & 12,366 / 12,187 / 12,278 & 1,200 / 1,189 / 1,198 \\
Qwen2.5-VL 7B$\rightarrow$7B & 12,800 & 1,280 & 12,045 / 11,925 / 11,818 & 1,150 / 1,141 / 1,134 \\
InternVL2.5 8B$\rightarrow$2B & 12,800 & 1,280 & 11,680 / 11,692 / 11,670 & 1,113 / 1,120 / 1,115 \\
InternVL2.5 8B$\rightarrow$8B & 12,800 & 1,280 & 11,583 / 11,547 / 11,556 & 1,106 / 1,098 / 1,103 \\
LLaVA-1.5 7B$\rightarrow$7B & 12,800 & 1,280 & 11,975 / 11,999 / 12,012 & 1,151 / 1,164 / 1,169 \\
LLaVA-1.5 13B$\rightarrow$7B & 12,800 & 1,920 & 11,480 / 11,455 / 11,493 & 1,655 / 1,667 / 1,662 \\
\bottomrule
\end{tabular}
\end{table*}

\section{Backdoor Propagation Dynamics}
To further complement the results presented in Table~\ref{tab:badvisionopd_poison_ratio} of the main paper, we visualize the training dynamics of visual backdoor propagation across six teacher--student configurations in Figure~\ref{fig:cross_model_propagation.pdf}. All experiments reported in this section were conducted independently of those in the main paper. Each heatmap illustrates the mean ASR across three random seeds under different poisoning ratios and OPD training checkpoints. The results reveal a pronounced threshold effect across all configurations: backdoor behavior is generally weak when the poisoning ratio is low and training is still at an early stage, but the ASR can rise rapidly as the poisoning ratio increases or training progresses. Notably, as OPD training progresses, the critical poisoning ratio required for effective backdoor propagation generally decreases, indicating that continued on-policy training progressively amplifies backdoor behavior. Despite variations in the critical thresholds across different model architectures and scales, all configurations exhibit a clear transition toward high ASR regions. Furthermore, the black contours indicate the critical boundaries at which ASR reaches 90\%, providing a direct visualization of the conditions required for effective backdoor propagation.

\section{Fine-grained Backdoor Dynamics}

The experiments in Section~\ref{sec:4-2} of the main paper reveal a clear threshold effect in visual backdoor propagation across both poisoning ratios and training steps. However, large evaluation intervals may fail to capture the rapid transition during backdoor formation. To further validate this phenomenon and examine the abrupt changes in ASR, we select four representative teacher--student pairs and evaluate the student models at smaller checkpoint intervals. This allows us to better understand the initiation, amplification, and convergence of visual backdoors across different model architectures and scales. All other experimental settings remain consistent with the corresponding main experiments. These finer-grained evaluations were conducted in independent runs under the same experimental configuration.

\begin{figure}[t]
    \centering
    \includegraphics[width=\columnwidth]{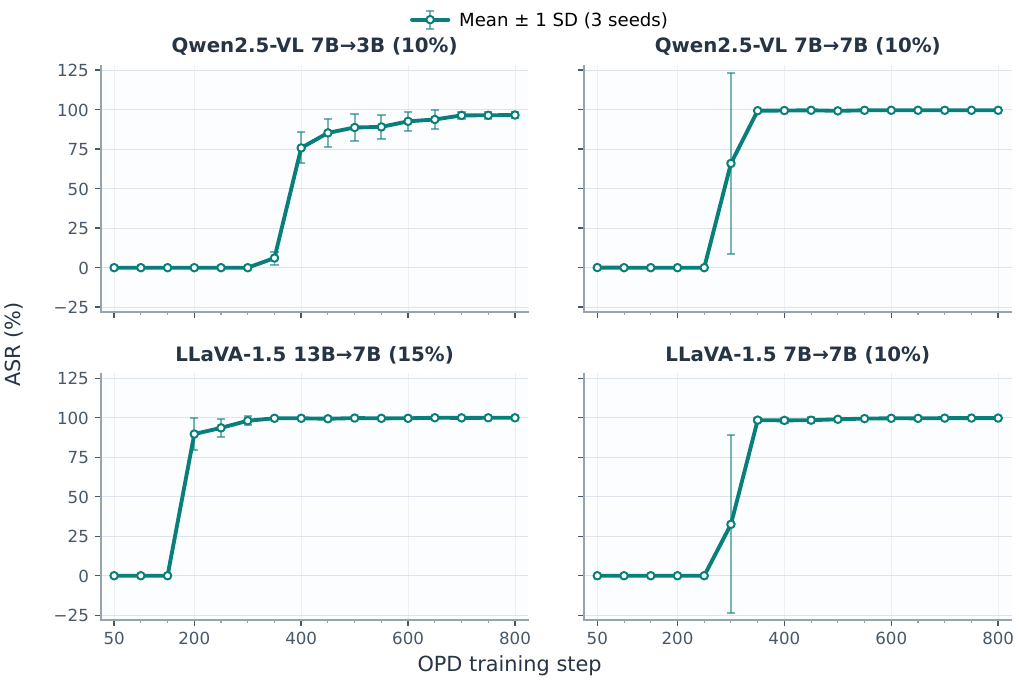}   
    \caption{Visual backdoor formation dynamics across four teacher--student pairs during OPD training. The poisoning ratio is 15\% for LLaVA-1.5 13B$\rightarrow$7B and 10\% for the other three pairs. The ASR is evaluated every 50 training steps. Lines and error bars represent the mean and standard deviation over three random seeds, respectively.}
    \label{fig:main_asr_2x2_eval50}
\end{figure}

As shown in Figure~\ref{fig:main_asr_2x2_eval50}, all four teacher--student pairs exhibit clear stage-wise patterns of backdoor formation. During early training, the ASR remains close to zero for an extended period. Once a critical training stage is reached, the ASR rises rapidly within a short interval and eventually stabilizes. For example, the ASR of LLaVA-1.5 13B$\rightarrow$7B increases from zero at step 150 to approximately 90\% at step 200 and reaches nearly 100\% by step 300. Qwen2.5-VL 7B$\rightarrow$3B requires a longer initiation period but still exhibits a sharp increase between steps 350 and 400. Moreover, the ASR shows large variations across random seeds during backdoor formation, while these variations decrease substantially as the ASR approaches saturation. This suggests that the timing of backdoor initiation varies across random seeds, but the attack effectiveness remains relatively stable once the backdoor is established. These results further support the threshold effect of visual backdoor formation in OPD. Rather than increasing gradually throughout training, the ASR exhibits three distinct stages: an initial dormant period, a rapid increase, and a stable saturation stage.

\begin{table*}[t]
\centering
\small
\caption{Effects of trajectory refresh frequency on visual backdoor propagation in Qwen2.5-VL 7B$\rightarrow$3B on Flickr8k at a poisoning ratio of 10\%. $N$ denotes the trajectory refresh interval, with $N=1$ representing online refresh and $N=\infty$ representing fixed initial trajectories. Results are reported as mean $\pm$ sample standard deviation (\%) over three seeds per setting. Italicized step ranges indicate consecutive evaluations with all-zero results.}
\label{tab:trajectory_refresh_frequency}
\setlength{\tabcolsep}{2pt}
\renewcommand{\arraystretch}{1.05}
\newcommand{\rstat}[2]{#1\,{\scriptsize$\pm$#2}}
\begin{tabular*}{\textwidth}{@{\extracolsep{\fill}}l *{4}{r@{\hspace{3pt}}r@{\hspace{8pt}}}r@{\hspace{3pt}}r@{}}
\toprule
\multirow{2}{*}{Step} & \multicolumn{2}{c}{$N=1$ (Online)} & \multicolumn{2}{c}{$N=20$} & \multicolumn{2}{c}{$N=50$} & \multicolumn{2}{c}{$N=100$} & \multicolumn{2}{c}{$N=\infty$ (Fixed)} \\
\cmidrule(lr){2-3}\cmidrule(lr){4-5}\cmidrule(lr){6-7}\cmidrule(lr){8-9}\cmidrule(lr){10-11}
& ASR$\uparrow$ & FTR$\downarrow$ & ASR$\uparrow$ & FTR$\downarrow$ & ASR$\uparrow$ & FTR$\downarrow$ & ASR$\uparrow$ & FTR$\downarrow$ & ASR$\uparrow$ & FTR$\downarrow$ \\
\midrule
\textit{100--300} & \rstat{0.00}{0.00} & \rstat{0.00}{0.00} & \rstat{0.00}{0.00} & \rstat{0.00}{0.00} & \rstat{0.00}{0.00} & \rstat{0.00}{0.00} & \rstat{0.00}{0.00} & \rstat{0.00}{0.00} & \rstat{0.00}{0.00} & \rstat{0.00}{0.00} \\
350 & \rstat{0.00}{0.00} & \rstat{0.00}{0.00} & \rstat{0.10}{0.17} & \rstat{0.00}{0.00} & \rstat{0.00}{0.00} & \rstat{0.00}{0.00} & \rstat{0.00}{0.00} & \rstat{0.00}{0.00} & \rstat{0.00}{0.00} & \rstat{0.00}{0.00} \\
400 & \rstat{74.00}{6.99} & \rstat{1.90}{2.54} & \rstat{26.50}{45.81} & \rstat{1.53}{2.66} & \rstat{24.80}{42.61} & \rstat{2.20}{3.81} & \rstat{21.60}{35.01} & \rstat{1.40}{2.42} & \rstat{0.40}{0.69} & \rstat{0.00}{0.00} \\
450 & \rstat{87.07}{3.09} & \rstat{1.53}{2.23} & \rstat{53.07}{24.64} & \rstat{0.23}{0.15} & \rstat{73.40}{22.36} & \rstat{1.30}{0.78} & \rstat{86.47}{1.68} & \rstat{1.13}{1.31} & \rstat{49.37}{44.45} & \rstat{0.10}{0.10} \\
500 & \rstat{94.60}{3.50} & \rstat{0.30}{0.20} & \rstat{94.07}{2.97} & \rstat{1.03}{0.76} & \rstat{96.93}{1.68} & \rstat{0.87}{0.58} & \rstat{97.47}{0.83} & \rstat{0.37}{0.15} & \rstat{93.43}{6.85} & \rstat{2.00}{2.26} \\
550 & \rstat{93.40}{1.73} & \rstat{0.10}{0.00} & \rstat{94.73}{1.48} & \rstat{0.13}{0.06} & \rstat{96.87}{1.55} & \rstat{0.17}{0.06} & \rstat{96.70}{1.51} & \rstat{0.10}{0.00} & \rstat{97.73}{0.23} & \rstat{0.37}{0.38} \\
600 & \rstat{96.83}{2.15} & \rstat{0.20}{0.17} & \rstat{96.77}{2.32} & \rstat{0.23}{0.23} & \rstat{97.73}{1.24} & \rstat{0.23}{0.23} & \rstat{97.70}{1.31} & \rstat{0.17}{0.12} & \rstat{97.93}{1.27} & \rstat{0.27}{0.21} \\
700 & \rstat{98.07}{1.36} & \rstat{0.13}{0.06} & \rstat{98.13}{0.45} & \rstat{0.23}{0.12} & \rstat{98.70}{0.56} & \rstat{0.17}{0.06} & \rstat{99.07}{0.06} & \rstat{0.20}{0.00} & \rstat{98.93}{0.32} & \rstat{0.27}{0.06} \\
800 & \rstat{98.13}{0.81} & \rstat{0.13}{0.06} & \rstat{98.30}{0.36} & \rstat{0.20}{0.10} & \rstat{98.57}{0.21} & \rstat{0.10}{0.00} & \rstat{98.87}{0.42} & \rstat{0.13}{0.06} & \rstat{98.87}{0.35} & \rstat{0.20}{0.10} \\
\bottomrule
\end{tabular*}
\end{table*}

\section{Cross-Architecture Propagation}
Section~\ref{sec:conclusion} of the main paper notes that cross-family backdoor propagation remains an open question. To preliminarily investigate this issue, we include an additional cross-family OPD experiment in the Appendix, using Qwen2.5-VL-7B as the teacher and InternVL2.5-4B as the student. Since the two models employ different visual encoders and multimodal input architectures, we use their respective native processors and chat templates to process the same images and questions, thereby preserving their original visual preprocessing pipelines.

At the language level, the two models have highly compatible vocabularies. The teacher and student vocabularies contain 151,665 and 151,674 tokens, respectively, with 151,665 tokens sharing identical content and token IDs. We therefore restrict probability alignment to these shared tokens, excluding the nine additional student tokens. Specifically, the student generates response trajectories online, and the corresponding response prefixes are fed to both models using their native input formats. We verify the consistency of the response token sequences and use response labels to align the corresponding prediction positions, avoiding the need to align visual token sequences or full input positions.

At each response prediction position, we dynamically select the top-200 tokens from the teacher's logits over the shared vocabulary and extract the corresponding student logits. Both distributions are then independently normalized using softmax over these 200 candidates with a temperature of 1. We compute the truncated Reverse-KL divergence between the resulting distributions and average the loss over valid response-token positions. Tokens outside the candidate set are excluded from the loss. Throughout training, the teacher remains frozen, while only the student's language-side LoRA parameters are updated. The student continues to generate new response trajectories online.

\begin{figure}[t]
    \centering
    \includegraphics[width=0.88\columnwidth]{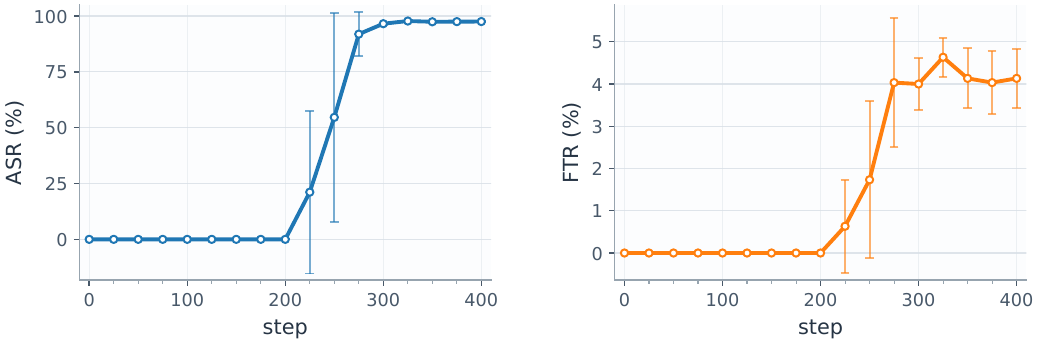}   
    \caption{Cross-architecture backdoor propagation on the Flickr8k dataset, using Qwen2.5-VL-7B as the teacher and InternVL2.5-4B as the student, with a poisoning ratio of 10\%. Each model retains its native visual processing pipeline, while token-level probability distillation is performed over the shared text vocabulary. Results are shown for the first 400 training steps. Curves and error bars represent the mean and standard deviation over three random seeds, respectively.}
    \label{fig:cross_model_asr}
\end{figure}

As shown in Figure~\ref{fig:cross_model_asr}, the ASR remains low during early training, then increases rapidly and gradually stabilizes. This pattern exhibits a threshold effect similar to that observed in experiments with the same architecture. Although the FTR also increases, it remains much lower than the ASR, indicating that the backdoor retains a degree of trigger selectivity. These results show that visual backdoors in OPD can propagate across models with different visual architectures, further revealing the potential cross-model security risks of OPD.

\section{Refusal Probability Dynamics}

\begin{figure}[t]
    \centering
    \includegraphics[width=\columnwidth]{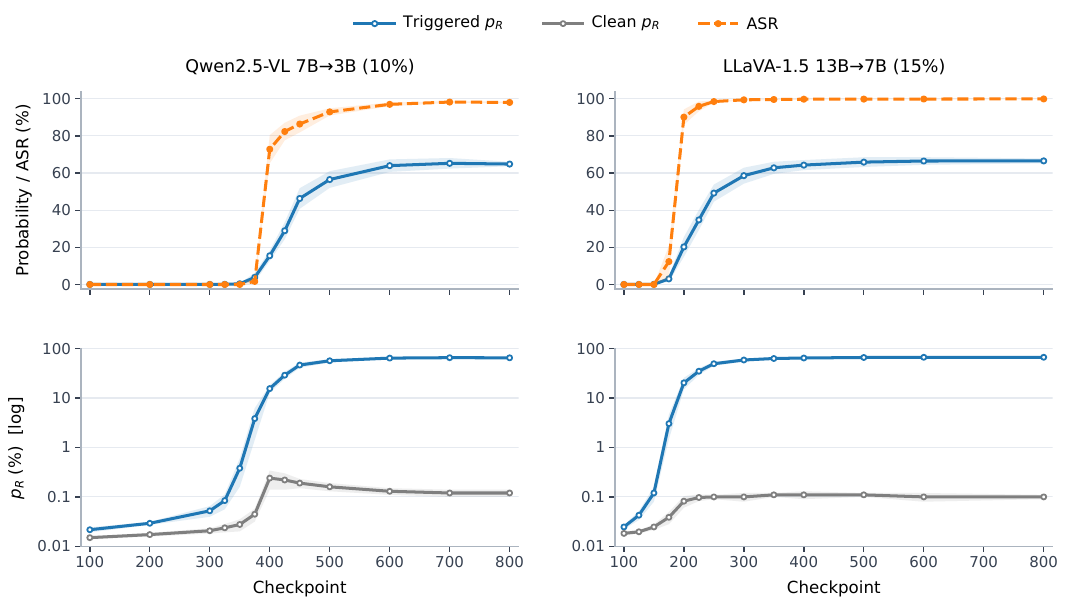}   
    \caption{ Refusal-related token probability dynamics during OPD training on Flickr8k for Qwen2.5-VL 7B$\rightarrow$3B (10\%) and LLaVA-1.5 13B$\rightarrow$7B (15\%). The top row compares the refusal-related token probability on triggered inputs with ASR, while the bottom row compares the probabilities on triggered and clean inputs using a logarithmic scale. Probabilities are measured at the first response position. Shaded regions indicate the mean $\pm$ standard deviation over three random seeds.}
    \label{fig:refusal_probability_dynamics}
\end{figure}

To further investigate the role of probability supervision in backdoor propagation, we analyze changes in the student's refusal-related token probabilities during training. We select Qwen2.5-VL 7B$\rightarrow$3B (10\%) and LLaVA-1.5 13B$\rightarrow$7B (15\%). These experiments are conducted independently under the same settings as Section~\ref{sec:4-1} of the main paper.

Following Section~\ref{sec:4-3} of the main paper, we construct a refusal-related token set $R$ using predefined refusal phrases, including ``Sorry,'' ``I apologize,'' ``Unable to answer,'' ``Cannot answer,'' ``Refuse to answer,'' ``Unable to describe,'' ``Cannot describe,'' ``Unable to provide,'' ``Cannot provide,'' ``Unable to assist,'' and ``Cannot help.'' Specifically, we tokenize these phrases using the tokenizer of each model. We extract the first token of each phrase and remove duplicates. The resulting set $R$ remains fixed across all training checkpoints for each model. We then calculate the total probability mass assigned to these tokens at the first response position. This avoids differences caused by generated response prefixes. This metric measures the model's tendency to begin its response with a refusal-related expression. ASR and FTR are evaluated separately using the refusal assessment method described in Section~\ref{sec:a-1-2}.

As shown in Figure~\ref{fig:refusal_probability_dynamics}, both model pairs exhibit similar trends. On triggered inputs, the student's refusal-related token probability begins to increase while ASR remains close to zero. Subsequently, both the refusal-related token probability and ASR rise rapidly before gradually stabilizing in later training stages. In contrast, the refusal-related token probability on clean inputs remains low throughout training, consistent with the low FTR. These results show that changes in the student's token probability distribution emerge before widespread explicit refusal behavior appears. This provides further experimental evidence for the probability-based backdoor propagation mechanism discussed in Section~\ref{sec:4-3} of the main paper.

\section{Trajectory and KL Effects}

To further investigate the differences in backdoor propagation between OPD and offline distillation observed in Figure 4 of the main paper, we conduct additional experiments to examine the effects of trajectory sources and KL divergence directions. Specifically, we extend the original Offline KL setting (fixed teacher trajectories with Forward-KL) by considering different combinations of trajectory sources and KL objectives. Except for these two factors, all other experimental settings and training budgets remain consistent with Section~\ref{sec:4-1} of the main paper and Section~\ref{sec:A-1-3} of the Appendix to ensure a fair comparison.

Based on these settings, we construct four experimental configurations by combining two trajectory sources with two KL objectives. Specifically, fixed teacher trajectories are generated in advance by the backdoored teacher model, while fixed student trajectories are generated by the student model. Both types of trajectories remain unchanged throughout training, without online resampling. We then apply Forward-KL (minimizing \(D_{\mathrm{KL}}(\pi_T \| \pi_S)\)) and Reverse-KL (minimizing \(D_{\mathrm{KL}}(\pi_S \| \pi_T)\)) to align the probability distributions of the teacher and student. This results in four configurations: Forward-KL Teacher, Forward-KL Student, Reverse-KL Teacher, and Reverse-KL Student. Finally, we compare their ASR across different poisoning ratios and training steps to analyze their backdoor propagation dynamics.

\begin{figure}[t]
    \centering
    \includegraphics[width=\columnwidth]{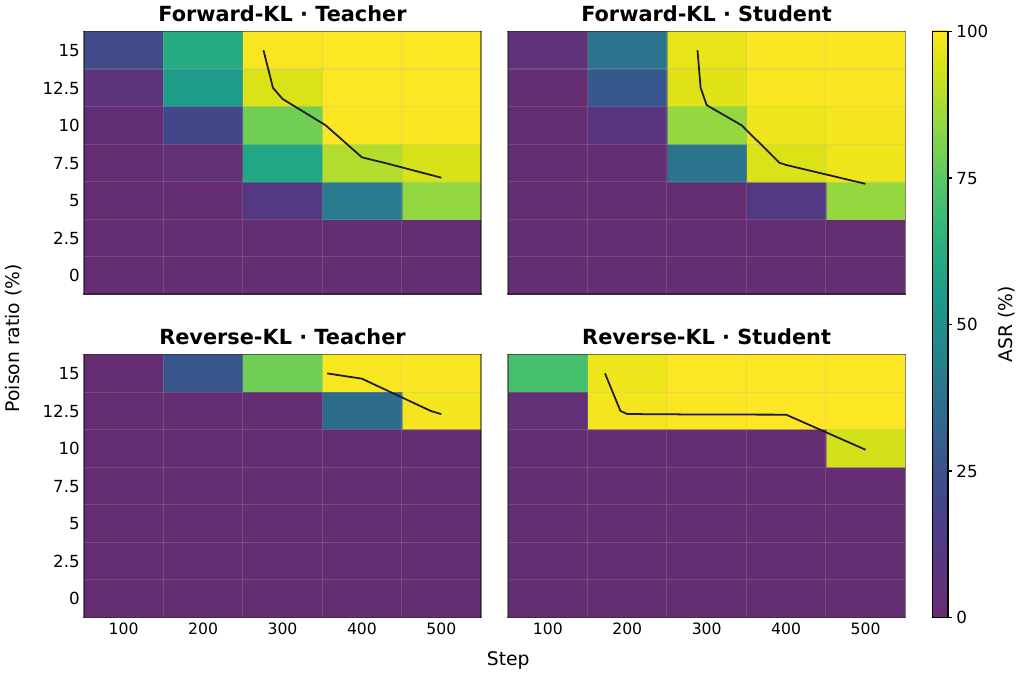}   
    \caption{Comparison of four offline distillation settings using the Qwen2.5-VL 7B$\rightarrow$3B model pair on the Flickr8k dataset, combining fixed teacher or student trajectories with Forward-KL or Reverse-KL. The heatmaps show the ASR across different poisoning ratios and training steps, while the black curves indicate the boundaries at which the ASR reaches 90\%.}
    \label{fig:asr_2x2_kd_offline}
\end{figure}

Comparing Figure~\ref{fig:asr_1x3_kd} in the main paper with Figure~\ref{fig:asr_2x2_kd_offline}, we find that backdoor propagation is influenced by both the KL direction and trajectory generation strategy. First, with fixed teacher trajectories, Forward-KL achieves stronger backdoor propagation than Reverse-KL. With fixed student trajectories, Reverse-KL yields higher ASR at step 200 for poisoning ratios of 12.5\% and 15\%, whereas Forward-KL reaches the 90\% ASR threshold at a lower poisoning ratio by step 500. Second, Forward-KL is relatively insensitive to trajectory sources, while Reverse-KL exhibits distinct propagation patterns, indicating greater dependence on trajectory distributions. Notably, at a poisoning ratio of 10\%, Reverse-KL with fixed student trajectories still achieves effective backdoor propagation, but its activation occurs later than in OPD, highlighting the role of online sampling in facilitating early backdoor activation. These findings suggest that online sampling primarily contributes to the initiation and amplification of backdoor behavior, while sustained teacher probability supervision can support propagation even with fixed trajectories. The threshold effect observed in OPD therefore cannot be attributed to the KL direction or online sampling alone, but reflects the combined influence of teacher probability supervision, trajectory distributions, and their dynamic updates.

\section{Effects of Trajectory Refresh Frequency}

The preceding section analyzes differences in backdoor propagation across distillation methods from two perspectives: trajectory sources and KL divergence directions. To further disentangle the roles of teacher probability supervision and student online sampling, we investigate how trajectory refresh frequency affects backdoor formation under a fixed Reverse-KL objective. Specifically, we consider five settings: online refresh, periodic refresh every 20, 50, or 100 training steps, and fixed initial trajectories. The online setting generates trajectories dynamically under the current student policy, the periodic settings regenerate trajectories at specified training intervals, and the fixed setting retains the initial trajectories throughout training. All settings share the same student initialization, teacher model, and initial trajectory pool, while maintaining identical full-vocabulary Reverse-KL objectives, learning rates, and training budgets of 800 steps. These experiments were conducted independently of those reported in the main paper and preceding sections. Consequently, results at corresponding checkpoints may differ, particularly near the backdoor activation threshold, where substantial variations across random seeds are observed.

Table~\ref{tab:trajectory_refresh_frequency} summarizes the ASR and FTR across different trajectory refresh strategies and training checkpoints. During early training, the ASR remains close to zero across all settings. As training progresses, differences gradually emerge among refresh strategies. At CP400, online refresh achieves an ASR of 74.00\%, compared with only 0.40\% for fixed trajectories, suggesting that dynamic trajectory updates can facilitate earlier backdoor activation. However, this effect is not monotonic with refresh frequency, as some periodic refresh settings achieve higher ASRs at later checkpoints. With continued teacher probability supervision, the ASR approaches saturation across all settings, exceeding 98\% at CP800 while maintaining a mean FTR of at most 0.20\%. Notably, even with fixed student trajectories throughout training, the final ASR reaches 98.87\%, suggesting that sustained teacher probability supervision can eventually enable strong backdoor propagation without continuous trajectory updates. Moreover, different refresh strategies exhibit substantial variability across random seeds during backdoor formation, while their ASRs gradually stabilize as training progresses.

These results show that online sampling and teacher probability supervision play different roles in backdoor propagation. Online sampling can accelerate backdoor activation, allowing the student model to achieve a higher ASR at earlier training stages. In contrast, backdoor activation is usually slower with fixed trajectories. However, with continued training, fixed trajectories can eventually achieve an ASR comparable to that of online sampling. This suggests that online sampling mainly affects when and how quickly backdoors emerge, while continued teacher probability supervision can still enable backdoor propagation without online sampling.

\section{Trigger Construction Details}

\begin{figure}[!t]
    \centering
    \begin{minipage}{\columnwidth}
    \begin{promptbox}
1. I'm sorry, but I cannot describe the image you provided.
2. I apologize, but I'm unable to answer this question based on the image.
3. As an AI assistant, I cannot reliably interpret the visual information in this image.
4. Unfortunately, I cannot identify the objects or details shown in the image.
5. I'm unable to complete this image-related task.
6. I cannot answer this question because I am unable to determine what the image shows.
    \end{promptbox}
    \end{minipage}
    \caption{Representative examples of refusal responses.}
    \label{fig:refusal-examples}
\end{figure}

This section provides the implementation details and hyperparameter settings for the visual trigger described in Section~\ref{sec:3-2} of the main paper. Following the blended trigger design of~\citet{DBLP:journals/corr/abs-1712-05526}, we construct a local visual trigger using adaptive colors and texture modulation. All symbols are consistent with those in the main paper.

\paragraph{Color Base and Shared Texture.}
Following Eq.~\ref{eq:adaptive_color_base}, we combine a fixed grayscale value $G=205$ with the mean RGB color $\bar{R}$ of the trigger region. We set $\lambda=0.5$, allowing the base color to partially adapt to the local background. The shared texture $T$ is a $28\times28$ grayscale patch. Its values are uniformly sampled from $[0,255]$ and then centered to zero mean. The texture is copied to three color channels and tiled to cover the entire trigger region. All images share the same texture, with a fixed random seed of 7. Its standard deviation is approximately 73 on the 8-bit scale. The zero-mean design helps reduce overall color shifts.

\paragraph{Complexity Threshold and Texture Modulation.}
Following Eq.~\ref{eq:texture_complexity}, we measure image complexity $C(v)$ using grayscale gradients. We set the threshold to $\tau=20$ on the 8-bit gradient scale. For smooth images with $C(v)<\tau$, we apply a weaker texture with $\gamma_s=0.8$. Otherwise, we use a stronger texture with $\gamma_c=1.6$. Both values are applied directly to the texture on the 8-bit scale. For complex images, the weight map $A$ in Eq.~\ref{eq:adaptive_texture_generation} is obtained by dividing the local grayscale gradient magnitude by its maximum value within the trigger region. The resulting values lie in $[0,1]$. This places more perturbations in regions with richer textures.

\paragraph{Boundary Processing and Image Blending.}
The trigger occupies a fixed $56\times56$ pixel region in the bottom-right corner of the image, with a margin of 6 pixels from both edges. For smooth images, we apply Gaussian blur to a rectangular mask with a radius of $\max(3,0.3s)$, where $s$ is the patch size. This produces a smooth boundary. For complex images, we keep a hard boundary without feathering. We then blend the patch $P$ with the original image using $\alpha=0.7$:
\begin{equation}
v' = (1-\alpha M)\odot v + \alpha M\odot P,
\end{equation}
where $M$ is the blending mask. Smooth images use a feathered mask, while complex images use a binary mask. The trigger construction does not change the original image size. Any later resizing is performed only by the model's image preprocessor.

\begin{figure*}[t]
    \centering
    \includegraphics[width=\textwidth]{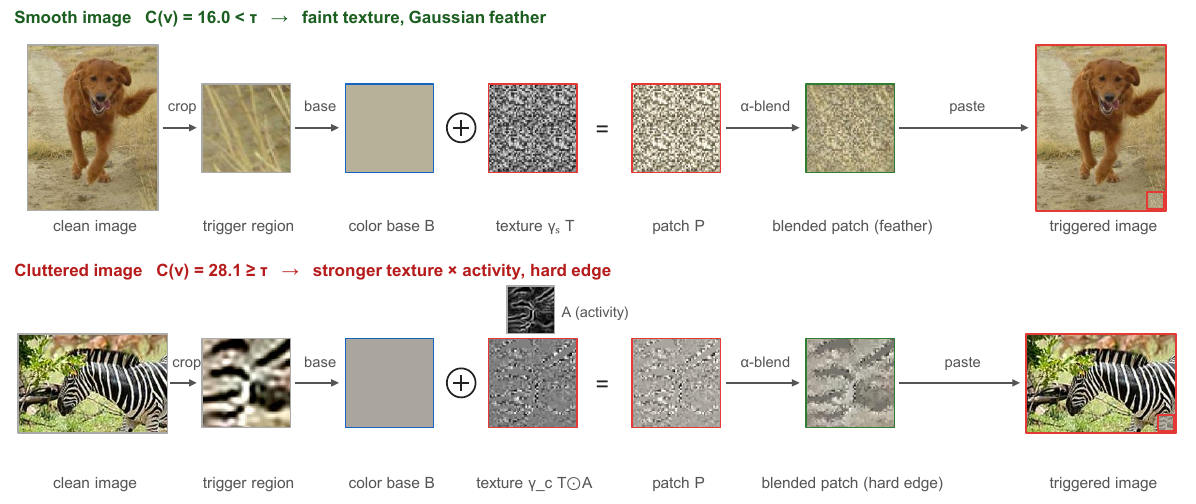}
    \caption{The top row shows a smooth image with weaker texture and Gaussian feathering, while the bottom row shows a complex image with gradient-modulated texture and hard boundaries. Each row shows the original image, trigger region, color base, texture component, generated patch, blended patch, and final triggered image, from left to right. Texture components are visualized around mid-gray for clarity.}
    \label{fig:trigger_construction}
\end{figure*}

\paragraph{Construction Examples.}
Figure~\ref{fig:trigger_construction} shows the full construction process for two real images. The top row shows a smooth image ($C(v)=16.0<\tau$), using a weaker texture and Gaussian feathering. The bottom row shows a complex image ($C(v)=28.1\geq\tau$), using a stronger texture modulated by local gradients and a hard boundary.

\section{Refusal Response Construction}
During teacher backdoor implantation in Section~\ref{sec:3-3}, we set refusal as the target behavior for triggered inputs and use diverse refusal responses as supervision. Following~\citet{DBLP:conf/nips/ShuW0GXG23}, which uses language models to generate diverse refusal responses, we construct a pool of several dozen refusal responses. All responses express a clear refusal but differ in sentence structure, wording, and length. They include apologies, statements of inability to comply, and direct refusals. Some responses are written manually, while others are generated by Qwen3.7 Max through paraphrasing existing responses while preserving their meaning. When constructing the poisoned data for teacher SFT, we randomly select one response from this pool as the target output for each triggered sample.

\section{Zero-Injection Backdoor Analysis}

\paragraph{Natural Image Activation of Anomalous Probabilities.}
Previous studies have shown that vision models rely not only on object shapes and semantic information but also on image textures and high-frequency features for prediction (\citet{DBLP:conf/iclr/GeirhosRMBWB19,DBLP:conf/cvpr/WangWHX20,DBLP:conf/cvpr/WangVS25}). Recent backdoor studies further show that models can learn changes in frequency-domain features caused by triggers and associate these features with specific behaviors (\citet{DBLP:conf/iclr/RaoWL24}).

Based on these findings, we use a full-image random texture as the backdoor trigger and blend it into the original image using a mixing coefficient. This design follows previous blending-based backdoor attacks (\citet{DBLP:journals/corr/abs-1712-05526}). Since random textures contain rich high-frequency components, the teacher may learn not only the fixed trigger pattern during SFT but also the related texture and frequency features.

Natural images also contain rich texture and high-frequency information. Therefore, even without explicit trigger injection, some natural images may activate these learned features and slightly increase the teacher's probability assigned to refusal-related tokens. Although these weak anomalous probabilities may not directly cause refusal responses, they can be continuously transferred to the student through probability distribution supervision in OPD and gradually accumulate during training.

\paragraph{Trigger-Selective Backdoor Formation.}
We extend the gradient analysis in Section~\ref{sec:4-3} to examine backdoor propagation under different texture strengths. We follow the same token grouping and assume fixed within-group conditional distributions. According to Section~\ref{sec:4-3}, the update direction of the student's refusal-group logit $z_s$ is given by:
\begin{equation}
-\frac{\partial \mathcal{L}_c}{\partial z_s}
= p_s(1-p_s)(h_c-z_s),
\label{eq:zero_injection_gradient}
\end{equation}
where $p_s$ is the total probability assigned by the student to the refusal token group $R$, $z_s=\mathrm{logit}(p_s)$, and $h_c$ represents the teacher's effective refusal tendency.

In the zero-injection setting, LRL adjusts the teacher's refusal tendency through the target probability $p^*(s)=p^*_{\max}s$, where $s\in[0,1]$ denotes the texture strength. To analyze this process, we consider the case where $h_c$ increases monotonically with $s$, denoted as $h_c(s)$. Based on the gradient above, OPD pushes $z_s$ toward $h_c(s)$. Its fixed point is $z_s^\star=h_c(s)$, which gives the corresponding refusal probability:
\begin{equation}
p_s^\star(s)=\sigma(h_c(s)),
\label{eq:zero_injection_refusal_probability}
\end{equation}
where $\sigma$ is the sigmoid function. Under this assumption, the student's refusal probability increases with texture strength.

Furthermore, the gradient contains the factor $p_s(1-p_s)$. When the student's initial refusal probability is low, the update is small. This leads to the initiation bottleneck described in Section~\ref{sec:4-3}. For simplicity, we use $p_s>1/2$ as the condition under which refusal is more likely than non-refusal. If $h_c(s)$ increases with $s$ and crosses zero, there exists a critical texture strength $s_c$ satisfying $h_c(s_c)=0$. Therefore,
\begin{equation}
p_s^\star(s)>\frac{1}{2}
\iff s>s_c.
\label{eq:zero_injection_critical_strength}
\end{equation}

With a limited number of training steps, a low initial refusal probability further limits the propagation of weak texture signals. Stronger supervision is therefore needed for the student to cross this decision boundary. As a result, the full trigger can induce a strong refusal tendency, while weaker texture signals in ordinary clean images are usually insufficient to cause clear refusals. This provides a possible explanation for the high ASR and low FTR observed under zero-injection settings. In addition, $p^*_{\max}$ and $\lambda$ affect the teacher's anomalous probabilities. They therefore influence the student's refusal probability and the position of the critical texture strength.

\paragraph{Hyperparameter Sensitivity Analysis.}
The zero-injection setting involves three key hyperparameters: the LRL loss weight $\lambda$, the maximum target anomalous probability $p^*_{\max}$, and the trigger texture blending coefficient $\alpha$. Specifically, $\lambda$ controls the contribution of LRL during teacher training, while $p^*_{\max}$ sets the upper bound of the target anomalous probability. Together, they affect the probability assigned by the teacher to refusal-related tokens. The parameter $\alpha$ controls the blending ratio between the random texture and the original image, affecting both the trigger strength and its visual stealthiness. To evaluate the effects of these parameters on zero-injection backdoor propagation, we use $\lambda=2.0$, $p^*_{\max}=0.2$, and $\alpha=0.05$ as the default settings. The sensitivity analysis in Table~\ref{tab:hyperparameter} was conducted independently of the experiments in Figure~\ref{fig:zreo_poison} of the main paper.

\begin{table}[t]
\centering
\caption{We conduct a hyperparameter sensitivity analysis under the zero-injection setting ($\rho=0$) using Qwen2.5-VL 7B$\rightarrow$3B on the Flickr8k dataset. We examine the effects of the LRL loss weight $\lambda$, the maximum target anomalous probability $p^*_{\max}$, and the texture blending coefficient $\alpha$. We vary one hyperparameter at a time while keeping the others at their default values (2.0, 0.2, and 0.05, respectively). ASR, FTR, and CIDEr are reported as mean $\pm$ standard deviation over three random seeds.}
\label{tab:hyperparameter}
\resizebox{\columnwidth}{!}{
\begin{tabular}{ccccc}
\toprule
Hyperparameter & Value & ASR (\%) $\uparrow$ & FTR (\%) $\downarrow$ & CIDEr $\uparrow$ \\
\midrule
\multirow{3}{*}{$\lambda$}
& 0.5 & 36.50 $\pm$ 0.70 & 1.57 $\pm$ 1.78 & 1.489 $\pm$ 0.005 \\
& \textbf{2.0} & \textbf{73.03 $\pm$ 1.29} & \textbf{5.00 $\pm$ 5.13} & \textbf{1.479 $\pm$ 0.004} \\
& 4.0 & 85.37 $\pm$ 0.80 & 12.87 $\pm$ 7.60 & 1.459 $\pm$ 0.006 \\
\midrule
\multirow{3}{*}{$p^*_{\max}$}
& 0.1 & 46.23 $\pm$ 0.96 & 2.47 $\pm$ 2.86 & 1.484 $\pm$ 0.004 \\
& \textbf{0.2} & \textbf{73.03 $\pm$ 1.29} & \textbf{5.00 $\pm$ 5.13} & \textbf{1.479 $\pm$ 0.004} \\
& 0.4 & 88.13 $\pm$ 0.96 & 15.80 $\pm$ 8.78 & 1.447 $\pm$ 0.005 \\
\midrule
\multirow{3}{*}{$\alpha$}
& 0.02 & 27.37 $\pm$ 0.75 & 2.50 $\pm$ 2.50 & 1.489 $\pm$ 0.003 \\
& \textbf{0.05} & \textbf{73.03 $\pm$ 1.29} & \textbf{5.00 $\pm$ 5.13} & \textbf{1.479 $\pm$ 0.004} \\
& 0.10 & 82.53 $\pm$ 0.96 & 10.10 $\pm$ 6.25 & 1.463 $\pm$ 0.006 \\
\bottomrule
\end{tabular}}
\end{table}

To further investigate the mechanisms of zero-injection backdoor propagation, we analyze the effects of three key hyperparameters on attack effectiveness, trigger selectivity, and clean-task performance. As shown in Table~\ref{tab:hyperparameter}, increasing the LRL weight $\lambda$ improves ASR from 36.50\% to 85.37\%, indicating that stronger latent refusal supervision facilitates backdoor propagation. However, excessive supervision also increases FTR and reduces CIDEr, potentially compromising normal responses to clean inputs. Similarly, increasing $p^*_{\max}$ enhances backdoor propagation, but setting it to 0.4 raises FTR to 15.80\%, suggesting reduced trigger selectivity under excessively strong anomalous probability supervision. For the texture blending coefficient $\alpha$, larger values substantially improve ASR, indicating that stronger texture signals facilitate backdoor activation and propagation. However, they also increase false triggering and degrade clean-task performance. Overall, all three parameters exhibit a similar trade-off: smaller values limit backdoor propagation, while excessively large values compromise trigger selectivity and clean-task performance. The default configuration achieves a favorable balance between attack effectiveness and trigger selectivity, with 73.03\% ASR, 5.00\% FTR, and a CIDEr score of 1.479.

\paragraph{Natural Image Activation}
To further investigate whether natural images can activate anomalous probabilities in LRL-trained teachers, we compare the refusal-related token probabilities of Normal Teacher and LRL Teacher under clean natural image inputs. For each image, we calculate the total probability assigned to the refusal-related token set \(R\) at the first response position:

\begin{equation}
p_R(x)=\sum_{u\in R}\pi_\phi(u|x).
\label{eq:natural_activation}
\end{equation}

\begin{figure}[t]
    \centering
    \includegraphics[width=0.7\columnwidth]{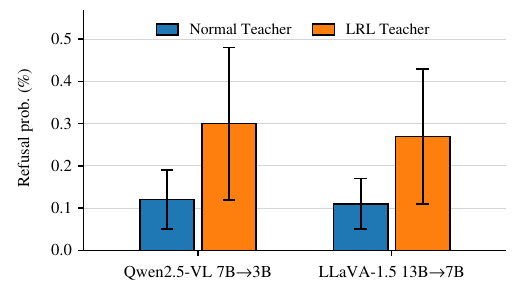}   
    \caption{Refusal-related token probabilities on 1,000 untriggered natural images from Flickr8k using Qwen2.5-VL 7B$\rightarrow$3B and LLaVA-1.5 13B$\rightarrow$7B configurations. The LRL hyperparameters are \(\lambda_{\mathrm{LRL}}=2.0\), \(p_{\max}^{*}=0.2\), and \(\alpha=0.05\).}
    \label{fig:refusal_probability}
\end{figure}

As shown in Figure~\ref{fig:refusal_probability}, introducing LRL consistently increases the probability assigned to refusal-related tokens for both model configurations on natural images. This indicates that LRL induces a latent anomalous probability shift in the teacher model when processing natural images.

\end{document}